\documentclass[journal,twoside]{IEEEtran}
\usepackage{framed, multirow}
\usepackage{amssymb}
\usepackage{latexsym}
\usepackage{url}
\usepackage{xcolor}
\usepackage{hyperref}
\definecolor{newcolor}{rgb}{.8,.349,.1}
\usepackage{tabularx}
\usepackage{threeparttable}
\usepackage{booktabs}
\usepackage[switch,columnwise]{lineno}
\usepackage{amsmath}
\usepackage{array}

\def\x{{\bf x}}

\def\0{{\bf 0}}
\def\1{{\bf 1}}

\def\etal{{\em et al.}}
\def\eg{{\em e.g.}}
\def\ie{{\em i.e.}}

\def\etal{{\em et al.\/}\,}

\usepackage{verbatim}
\usepackage{soul}
\usepackage{graphicx}
\usepackage{cite}

\begin{document}

%\verso{Fang \textit{et~al.}}

%\begin{frontmatter}

\title{Source-Free Collaborative Domain Adaptation via Multi-Perspective Feature Enrichment for Functional MRI Analysis}

\author{Yuqi~Fang, \and
Jinjian~Wu, \and
Qianqian~Wang, \and
Shijun~Qiu, \and 
Andrea Bozoki, \and 
Huaicheng Yan, \and
Mingxia~Liu% <-this % stops a space
\thanks{Y.~Fang, Q.~Wang and M.~Liu are with the Department of Radiology and Biomedical Research Imaging Center, University of North Carolina at Chapel Hill, Chapel Hill, NC 27599, USA. 
J.~Wu is with the Department of Acupuncture and Rehabilitation, The Affiliated Hospital of TCM of Guangzhou Medical University, Guangzhou, Guangdong, 510130, China.
S.~Qiu is with the Department of Radiology, The First Affiliated Hospital of Guangzhou University of Chinese Medicine, Guangzhou, Guangdong, 510120, China.
A.~Bozoki is with the Department of Neurology, University of North Carolina at Chapel Hill, NC 27599, USA. 
H.~Yan is with the Key Laboratory of Smart Manufacturing in Energy Chemical Process of Ministry of Education, East China University of Science and Technology, Shanghai, 200237, China. 
Corresponding author: M.~Liu (mxliu@med.unc.edu).}
}

\maketitle

\begin{abstract}
Resting-state functional MRI (rs-fMRI) is increasingly employed in multi-site research to aid neurological disorder analysis.
Existing studies usually suffer from significant cross-site/domain data heterogeneity caused by site effects such as differences in scanners/protocols.
Many methods have been proposed to reduce fMRI heterogeneity between source and target domains, heavily relying on the availability of source data.
But acquiring source data is challenging due to privacy concerns and/or data storage burdens in multi-site studies.
To this end, we design a \emph{source-free collaborative domain adaptation} (SCDA) framework for fMRI analysis, where \emph{only a pretrained source model and unlabeled target data are accessible}.
%Specifically, we first build a source model with labeled source data in a supervised manner.
Specifically, a multi-perspective feature enrichment method (MFE) is developed for target fMRI analysis, consisting of multiple collaborative branches to dynamically capture fMRI features of unlabeled target data from multiple views. 
Each branch has a data-feeding module, a spatiotemporal feature encoder, and a class predictor.
A mutual-consistency constraint is designed to encourage pair-wise consistency of latent features of the same input generated from these branches for robust representation learning. % of target data. 
To facilitate efficient cross-domain knowledge transfer without source data, we initialize  MFE using parameters of a pretrained source model. 
We also introduce an unsupervised pretraining strategy using 3,806 unlabeled fMRIs from three large-scale auxiliary databases, aiming to obtain a general feature encoder. 
%to further capture diverse and generalizable fMRI patterns. 
% (\ie, ABIDE, REST-meta-MDD, and ADHD-200).
Experimental results on three public datasets and one private dataset demonstrate the efficacy of our method in \emph{cross-scanner} and \emph{cross-study} prediction tasks. 
The model pretrained on large-scale rs-fMRI data has been released to the public. 
% The identified discriminative brain regions can be used as potential biomarkers to assist fMRI-based brain disorder analysis in clinical practice.
\end{abstract}

\begin{IEEEkeywords}
Source-free domain adaptation, feature enrichment, collaborative learning, unsupervised pretraining.
\end{IEEEkeywords}

% Autism Brain Imaging Data Exchange (ABIDE)
% REST-meta-MDD  Consortium (REST-meta-MDD)
% Attention-Deficit Hyperactivity Disorder 200 Consortium (ADHD-200)

\section{Introduction}
\IEEEPARstart{N}{eurological} disorders include a broad range of brain diseases, such as autism spectrum disorder and cognitive impairment~\cite{lord2020autism, van2018vascular}, caused by a combination of genetic, environmental, and psychological factors.
The treatments for these brain diseases emerge challenges to health care systems globally, bringing society much economic and manpower burden. 
Therefore, early and accurate identification of these disorders is much critical to helping clinicians develop timely interventions. 
Resting-state functional MRI (rs-fMRI) provides a non-invasive technique to measure spontaneous brain activity through blood-oxygen-level-dependent (BOLD) signals~\cite{fox2007spontaneous}. 
It has been increasingly used to study functional brain systems and has shown great clinical value in neurological disorder analysis~\cite{deshpande2015fully}.

\begin{figure}[!t]\centering
    \setlength{\belowcaptionskip}{-2pt}
    \setlength{\abovecaptionskip}{-2pt}
    \setlength\abovedisplayskip{-2pt}
    \setlength\belowdisplayskip{-2pt}
    \includegraphics[width=0.49\textwidth]{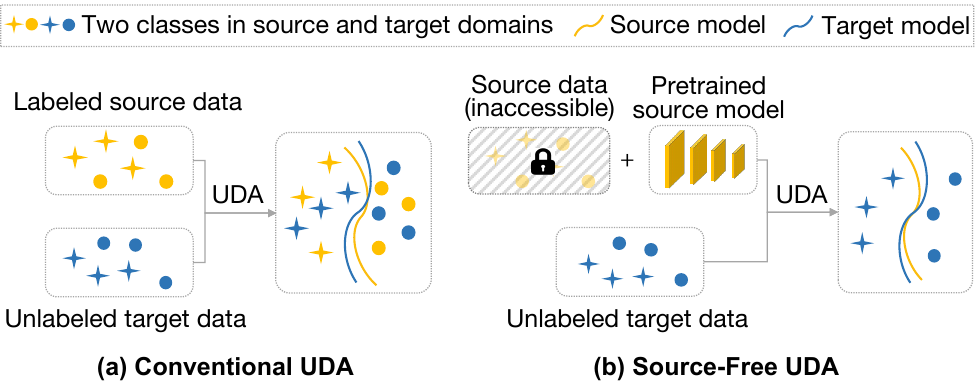}
    \caption{Illustration of (a) conventional unsupervised domain adaptation (UDA) and (b) source-free unsupervised domain adaptation (source-free UDA). Compared with conventional UDA that relies on labeled source data and unlabeled target data to reduce domain shift, source-free UDA only leverages a pretrained source model and unlabeled target data to learn a target model for improving adaptation performance, with source data inaccessible.}\label{fig_UDAvsSFUDA}    
\vspace{-12pt}
\end{figure}

Many studies~\cite{gallo2021thalamic, shi2021multivariate} employ multi-site fMRI to increase sample size for brain disorder identification, but many of them neglect \emph{significant cross-site/domain data heterogeneity} caused by site effects such as differences in scanner vendors and/or imaging  protocols~\cite{yamashita2019harmonization}, leading to suboptimal learning performance. 
Several recent studies~\cite{huang2020conditional, gao2019decoding} explore unsupervised domain adaptation (see Fig.~\ref{fig_UDAvsSFUDA}~(a)) to reduce fMRI heterogeneity between source and target domains. 
They generally heavily rely on the \emph{availability of source data} for target model construction. 
For instance, Gao~\etal~\cite{gao2019decoding} propose to use labeled source fMRI to generate a distinct descriptor, and then minimize the discrepancy between source and target distributions to achieve cross-subject decoding.
Huang~\etal~\cite{huang2020conditional} also rely on labeled source fMRI to learn a discriminative fMRI representation, and then they perform cross-domain adaptation using an adversarial-based neural network.
However, direct accessing source data is usually challenging in multi-site studies due to privacy concerns and/or data storage burdens.
Medical data may contain %patients' identifiable 
subjects' personal information, and directly accessing these data may cause privacy leakage~\cite{bateson2022source}.
Additionally, multi-site studies often have a large amount of data, which may cause a data storage burden~\cite{xia2021adaptive}. %, especially in infrastructure with insufficient memory~\cite{xia2021adaptive}.

%The aim of this study is to learn a target model for improving adaptation performance only based on a pretrained source model and unlabeled target fMRI data.

%This significantly limits their utility in practice due to privacy concerns and/or data storage burdens in multi-site studies.

To this end, we propose a novel \emph{source-free collaborative domain adaptation} (SCDA) framework for fMRI analysis and brain disorder diagnosis, %, where only a pretrained source model and unlabeled target data are accessible. 
without accessing any source data for target model training.
%Source-free unsupervised domain adaptation aims to adapt a pretrained source model to target domain(s) by only leveraging the inherent characteristics of unlabeled target data.
%The illustration of source-free unsupervised domain adaptation and conventional unsupervised domain adaptation is shown in Fig.~\ref{fig_UDAvsSFUDA}.
This study %will develop a source-free unsupervised domain adaptation method, 
aims to learn a target model for improving adaptation performance only based on a pretrained source model and unlabeled target fMRI data (see Fig.~\ref{fig_UDAvsSFUDA}~(b)).
Specifically, %we first build a source model with labeled source data in a supervised manner (afterward, source data are inaccessible).
we design a multi-perspective feature enrichment method (MFE), which consists of multiple collaborative branches to dynamically exploit unlabeled target data from multiple perspectives, \eg, window warping, receptive field manipulation, and window slicing.
Each branch of the MFE comprises a data-feeding module, a spatiotemporal feature encoder, and a class predictor. 
A mutual-consistency constraint is also designed to enforce pair-wise consistency of latent features generated by these branches for robust representation learning of unlabeled target data. 
To facilitate source-to-target knowledge transfer without accessing source data, we initialize the MFE using parameters of a pretrained source model. 
To obtain a general fMRI encoder, we further propose an effective unsupervised pretraining strategy based on 3,806 unlabeled fMRI data from three large-scale auxiliary fMRI databases, \ie, ABIDE~\cite{craddock2013neuro}, REST-meta-MDD~\cite{yan2019reduced}, and ADHD-200~\cite{bellec2017neuro}.
%This strategy enables our SCDA to capture a broad range of variations and complexities exhibited in fMRI data, facilitating effective cross-domain knowledge transfer. 
We evaluate our proposed SCDA on 1,591 rs-fMRI scans from three public datasets and one private dataset, with experimental results suggesting its efficacy in both cross-scanner and cross-study prediction tasks.
And the identified discriminative brain regions can be used as potential biomarkers to assist fMRI-based brain disorder analysis. % in clinical practice. 
%To the best of our knowledge, this is among the first attempts to investigate source-free domain adaptation for fMRI analysis. 
The contributions of this work are summarized below. 
\begin{itemize}
%\vspace{-1mm}
\item A novel source-free collaborative domain adaptation (SCDA) framework is designed for fMRI-based neurological disorder diagnosis without accessing source data. 
To our knowledge, this is among the first attempts to investigate source-free domain adaptation in fMRI analysis. 
%\vspace{1mm}
\item A multi-perspective feature enrichment method is designed to exploit dynamic characteristics of target data from multiple views. 
It can be straightforwardly applied to other applications for multi-view feature learning. %\eg, window warping, receptive field manipulation, and window slicing.  
%\vspace{1mm}
\item An unsupervised pretraining strategy is proposed to leverage large-scale auxiliary unlabeled fMRI databases to generate a general feature encoder, which helps learn more transferable fMRI features and enhances effective cross-domain adaptation. 
The model pretrained on large-scale rs-fMRI data has been shared with the public\footnote{\url{https://github.com/yqfang9199/SCDA}}. 
%\vspace{1mm}
\item Comprehensive experiments in cross-scanner and cross-study prediction tasks using three public datasets and one private dataset demonstrate the effectiveness of the SCDA. 
And the disorder-associated brain regions identified by the SCDA could be used as potential biomarkers to facilitate fMRI-based neurological disorder analysis. 
\end{itemize}

\vspace{-1mm}    
The remainder of this paper is organized as follows. 
Section~\ref{sec_relatedwork} reviews the most relevant studies on fMRI adaptation and fMRI representation learning.
Section~\ref{sec_method} introduces details of the proposed method.
Section~\ref{sec_experiment} presents materials and results.
Section~\ref{sec_discussion} discusses the influence of key components of our method.
The paper is concluded in Section~\ref{sec_conclusion}.

\begin{figure*}[!t]\centering
\setlength{\belowcaptionskip}{-2pt}
\setlength{\abovecaptionskip}{-2pt}
\setlength{\abovedisplayskip}{-2pt}
\setlength{\belowdisplayskip}{-2pt}
\includegraphics[width=1.0\textwidth]{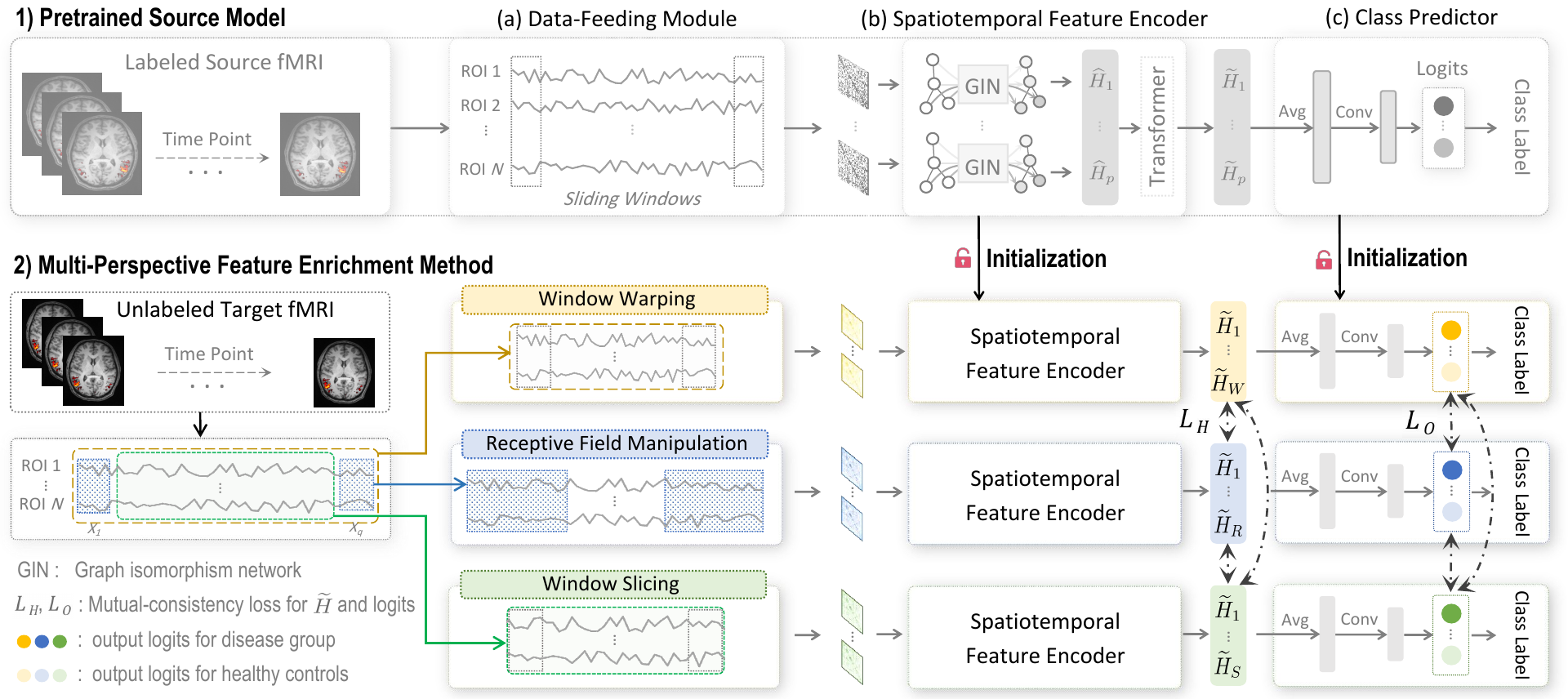}
\caption{Illustration of the proposed \emph{source-free collaborative domain adaptation} (SCDA) framework, which performs source-to-target domain adaptation only based on a pretrained source model and unlabeled target fMRI data.
Specifically, we first train a source model with labeled source data for extracting discriminative fMRI features.
Then, we design a multi-perspective feature enrichment method (MFE), which contains multiple collaborative branches to dynamically exploit unlabeled target data from various perspectives, \eg, window warping, receptive field manipulation, and window slicing.
Each branch is comprised of (a) a data-feeding module, (b) a spatiotemporal feature encoder, and (c) a class predictor.
To facilitate cross-domain knowledge transfer, we initialize the MFE using parameters of the pretrained source model.
Moreover, a mutual-consistency constraint $L_M=L_H+L_O$ is designed to enforce pair-wise consistency of latent features and output logits generated from these branches for robust representation learning of target fMRI.}\label{fig_pipeline}
\end{figure*}

\section{Related Work}\label{sec_relatedwork}
  
\subsection{Functional MRI Adaptation}
Functional MRI is increasingly employed to study brain systems and has shown great clinical value in neurological disorder diagnosis~\cite{zhao2022attention, cui2022braingb}.
To increase sample size and enhance statistical power, many studies~\cite{gallo2021thalamic, shi2021multivariate} propose to leverage multi-site/domain fMRI data for brain disorder analysis.
These studies usually assume that fMRI data are sampled from the same distribution while neglecting \emph{significant inter-domain heterogeneity} caused for instance by the use of different scanners/protocols, which could lead to biased or inaccurate predictions.
To address this domain distribution shift problem, many fMRI adaptation approaches~\cite{wang2019identifying, zhang2020transport, fang2023unsupervised, gao2019decoding} have been proposed to perform data/feature harmonization across domains, yielding robust prediction models to improve inference in target domain. 
Even with improved prediction performance, 
these studies heavily rely on the availability of source data for source-to-target adaptation. 
%For example, 
Zhang~\etal~\cite{zhang2020transport} present a multi-way adaptation architecture, which performs joint distribution alignment in each pair of source and target fMRI based on optimal transport. 
Gao~\etal~\cite{gao2019decoding} propose to use source fMRI to derive a distinct descriptor, and then they perform cross-subject adaptation by reducing discrepancy between source and target distributions.  
But acquiring source data is usually challenging due to privacy concerns and/or data storage burdens, while existing studies seldom investigate fMRI adaptation without accessing source data. 
%To the best of our knowledge, existing studies have not investigated cross-domain adaptation using fMRI data in source-free scenarios.
To bridge this gap, we propose a source-free collaborative domain adaptation method, where only a pretrained source model and unlabeled target data are available for target model construction. %, without accessing any source data.
We aim to transfer knowledge from a pretrained source model to unlabeled target data to improve target inference. %a target model and improve target inference by exploiting inherent structure and characteristics of unlabeled target data.

\subsection{Functional MRI Representation Learning}
% Existing studies have investigated fMRI feature representation learning for automated neurological disorder diagnosis.
Representation learning aims to extract informative features from fMRI data, which can facilitate subsequent fMRI analysis and interpretation of brain activity. 
Existing studies for fMRI representation learning can be mainly categorized into traditional machine learning methods and deep learning based methods. 
The traditional methods usually extract handcrafted fMRI features (\eg, clustering coefficient, modularity, and density) for downstream prediction~\cite{zhang2019classificatione}, and the commonly used prediction models include support vector machine (SVMs)~\cite{zhang2019classificatione}, eXtreme Gradient Boosting (XGBoost)~\cite{torlay2017machine}, and random forest (RF)~\cite{douglas2011performance}.
In these methods, the extracted human-engineered features require much expert knowledge, which may not be optimal for subsequent prediction.

To tackle the problem, many deep learning methods have been developed to automatically learn task-oriented fMRI features~\cite{zhang2022classification, li2021braingnn, kim2021learning, zhang2020multiview, wang2020multikernel}, which usually show superior performance compared with traditional machine learning models. 
For example, Li~\etal~\cite{li2021braingnn} propose an interpretable graph neural network, which leverages both topological and functional characteristics of fMRI data for neurological disorder diagnosis. 
% And this study can automatically identify salient brain regions, which have been verified related to task-specific brain community patterns. 
Kim~\etal~\cite{kim2021learning} introduce an attention-based neural network to incorporate both spatial and temporal information of fMRI data for brain connectome analysis.
Among existing fMRI studies, we find they usually leverage original fMRI timeseries directly, without considering multi-perspective representations of fMRI. % while not well exploiting fMRI data in a more comprehensive way. 
For example, some works~\cite{liu2023spatialt, luo2021shared} use sliding window strategy to characterize temporal dynamics of fMRI data, but they usually employ fixed window size, limiting the ability to capture brain temporal variability from various receptive fields.
%And many studies may neglect coarse-to-fine grained fMRI representations 
%they seldom consider performing window warping (\eg, upsampling and downsampling) of the original fMRI, and thus different fine-grained fMRI representations may be neglected.
%Also, current studies mainly utilize full-length fMRI timeseries and ignore slicing them into multiple shortened segments to augment sample size and improve model robustness. 
%Moreover, some works~\cite{liu2023spatialt, luo2021shared} use sliding window strategy to characterize temporal dynamics of fMRI data, but they usually employ fixed window size, limiting the ability to capture brain temporal variability from various receptive fields. 
In this work, we propose to dynamically exploit fMRI data from multiple perspectives, \eg, window warping, receptive field manipulation, and window slicing, which helps capture multi-view informative fMRI features.
Furthermore, our method introduces an unsupervised pretraining strategy, by training our model on 3,806 subjects from three large-scale auxiliary fMRI databases. 
Benefiting from these massive and diverse data, it is expected to capture more general fMRI features for improving performance in downstream tasks.

\section{Methodology}\label{sec_method}
% clear motivation on why this design is necessary; why it is working; source-free domain adaptation background is missing
To protect source data privacy and reduce data storage burden in multi-site studies, we develop a source-free collaborative domain adaptation (SCDA) framework for fMRI analysis, where \emph{only a pretrained source model and unlabeled target data are accessible} for target model construction.
As shown in Fig.~\ref{fig_pipeline}, the SCDA consists of 1) a pretrained source model for discriminative fMRI feature extraction, and 2) a multi-perspective feature enrichment method (MFE) for target inference using unlabeled target data from various views. 
The MFE is initialized using parameters of the pretrained source model for cross-domain knowledge transfer, and optimized via a mutual-consistency constraint.
We also propose an unsupervised pretraining strategy to obtain a general fMRI feature encoder, which aims to facilitate cross-domain adaptation.

\subsection{Construction of Pretrained Source Model}\label{sec_pretrain_source_model}

A pretrained source model is used to facilitate source-to-target knowledge transfer. %, we first need to obtain a pretrained source model based on labeled source data.
As shown in the top of Fig.~\ref{fig_pipeline}, the source model takes full-length fMRI timeseries as input and consists of \textit{a data-feeding module}, \textit{a spatiotemporal feature encoder}, and \textit{a class predictor}, elaborated as follows.

\subsubsection{Data-Feeding Module}\label{sec_datafeed_comp} 
In this module, each subject is represented by its fMRI timeseries $X \in \mathbb{R}^{L \times N}$, where $L$ and $N$ denote the number of time points and regions-of-interest (ROIs), respectively. 
To characterize brain functional variability across time, we partition $X$ into multiple overlapping sliding windows, and construct $p=\lfloor L-l/s\rfloor$ windowed timeseries $\{X_t \}_{t=1}^{p} \in \mathbb{R}^{l \times N}$ 
based on a sliding window with the length of $l=40$ and the stride of $s=30$, where $\lfloor.\rfloor$ is floor function.
Here, we represent each windowed timeseries $X_t$ as a functional brain network/graph, where graph nodes indicate brain ROIs and edges represent functional connections among these ROIs.  
% This practice has been verified by many brain studies~\cite{}. 
Denote $V_t$, $E_t$, and $A_t$ as the vertex set, the edge set, and the adjacency matrix, respectively.   
At the $t$-th sliding window, we represent $X_t$ as a graph $G_t = (V_t, E_t, A_t)$, where each node $v_{t,i}\in{V_t}$ represents the $i$-th ROI with its feature vector $h_{t,i}$ and each edge $(v_{t,i}, v_{t,j})\in E_t$ represents the Pearson's correlation coefficient (PCC) between the $i$-th and the $j$-th ROIs. 
The adjacency matrix $A_t\in\{0,1\}^{N\times N}$ is derived by keeping the top 30\% strongest edges of each graph, where $1$ denotes there exists an edge between two ROIs while $0$ denotes there is no edge. 

\subsubsection{Spatiotemporal Feature Encoder}\label{sec_feature_encoder}
With a set of graphs $\{G_t\}_{t=1}^{p}$ for each subject as input, a spatiotemporal feature encoder is designed to learn fMRI features from spatial and temporal aspects.
% Spatial information in rs-fMRI data indicates brain functional connectome, and effective extraction of these spatial features helps understand the node activity and coactivation pattern between connected ROIs~\cite{karahanouglu2015transient}.
% Temporal information models dynamic characteristics inherent in fMRI timeseries, which provides a deep understanding of neural activity and can serve as potential biomarkers for neurological disorders~\cite{zhao2019four}. 
Specifically, we introduce 1) a graph neural network for \emph{spatial feature aggregation} across ROIs and 2) a Transformer for \emph{temporal attention learning} across sliding windows, which are detailed as follows.

\textbf{\emph{1) Spatial Feature Aggregation.}}
Graph neural networks (GNNs) typically update node representations by aggregating features of their neighbors iteratively, thereby capturing spatial/structural information across the entire graph~\cite{li2021braingnn, bai2021twostream}.
We employ graph isomorphism network (GIN)~\cite{xu2018powerful} for spatial representation learning, a GNN variant with high discriminative power in graph classification.
Mathematically, the node-level embedding at the $t$-th sliding window, formulated as:
\begin{equation}\label{eq_spatial2}
h_{t,i}^k = MLP^k((1+\epsilon^k)\cdot h_{t,i}^{k-1}+\sum\nolimits_{u\in\mathcal{B}(i)}h_{t,u}^{k-1}),
\end{equation}
where $h_{t,i}^k\in\mathbb{R}^{D}$ denotes the feature vector of node $i$ at layer $k$, $h_{t,i}^0$ is one-hot encoding of node $i$, $MLP$ is a multiple-layer perceptron, $\epsilon$ is a learnable parameter initialized with 0, and $\mathcal{B}(i)$ denotes neighbors of node $i$.
Following~\cite{kim2021learning}, 
with {{$H_t^k=[h_{t,1}^k, \cdots, h_{t,N}^k]\in\mathbb{R}^{D\times N}$}}, 
we can formulate Eq.~\eqref{eq_spatial2} as:
\begin{equation}\label{eq_spatial3}
H_t^k = \sigma((\epsilon^k\cdot I+A_t)H_t^{k-1}W_t^k),
\end{equation}
where $H_t^k$ denotes all node features for the $k$-th layer at sliding window $t$, $I$ is an identity matrix, $A_t$ is an adjacent matrix, $W_t$ means $MLP$ weights, and $\sigma$ is a nonlinear function.
By concatenating node features of all $K$ layers, we obtain a new node-level embedding $H_t \in \mathbb{R}^{D\times N}$ at sliding window $t$.
%To characterize unique contributions of different brain regions to disease diagnosis,
We further embed a squeeze-excitation block~\cite{hu2018squeeze} into $H_t$ to automatically locate informative ROIs, where a spatial attention vector is represented as $M = sigmoid(W_2\psi(W_1\varphi(H_t))) \in\mathbb{R}^{N}$. 
Here, $\varphi$ calculates the mean of $H_t$, % along channel of brain ROIs
$W_1\in\mathbb{R}^{D\times D}$ and $W_2\in\mathbb{R}^{N\times D}$ are learnable weight matrices, $\psi$ is an activation function with batch normalization~\cite{ioffe2015batch} and Gaussian error linear unit~\cite{hendrycks2016gaussian}. 
Each element in $M$ denotes the importance of a specific ROI. 
The spatial-attended features $\hat{H}_t\in\mathbb{R}^{D}$ ($D=64$) can be obtained by multiplying $H_t$ and $M$, followed by an average operation across ROIs.

\textbf{\emph{2) Temporal Attention Learning.}}
With spatial-attended features ${\hat{H}}$ $=$ $[\hat{H}_1, \cdots, \hat{H}_p]$ of $p$ sliding windows for each subject, we also capture temporal attention across different windows. 
Denote $\phi_1$, $\phi_2$, and $\phi_3$ as linear operations. 
Specifically, we model such temporal attention using a single-head Transformer~\cite{vaswani2017attention} based on a self-attention matrix $Z=softmax(QK^T/\sqrt{d})$, where $Q=\phi_1(\hat{H})$, $K=\phi_2(\hat{H})$, and $d$ is a scaling factor for computation stability.
% With $Z$, one can tell which time windows contribute more to prediction.
Then, we can derive spatiotemporally-attended feature ${\tilde{H}}$ by multiplying $\phi_3(\hat{H})$ and $Z$, followed by an $MLP$ for feature abstraction. %to generate a whole-graph embedding at each time segment. %, $[\tilde{H}_1, \cdots, \tilde{H}_o]$.

\subsubsection{Class Predictor}\label{sec_class_predictor}
To obtain a whole-graph embedding for each subject, we average the spatiotemporal fMRI features ${\tilde{H}}$ across all sliding windows.
The derived embedding is fed into a linear fully-connected layer to output a prediction logit $O$. 
With the generated logits $O$ and ground-truth labels for all source samples, a cross-entropy loss function is used to train the source model in a supervised manner. 
In this way, we can obtain a pretrained source model for downstream target tasks, and \emph{the source data and labels are not available for target model construction}.

\subsection{Multi-Perspective Feature Enrichment} \label{sec_tribranch} 
To perform source-free domain adaptation with a pretrained source model and unlabeled target data, we propose a multi-perspective feature enrichment method (MFE). 
As shown in the bottom of Fig.~\ref{fig_pipeline}, it consists of multiple collaborative branches to dynamically capture target fMRI features from three views (\ie, window warping, receptive field manipulation, and window slicing).  
\if false
We initialize MFE using parameters of the pretrained source model to facilitate knowledge transfer. 
Moreover, a mutual-consistency constraint is designed to enforce pair-wise consistency of latent features and output logits generated from these branches for robust representation learning of unlabeled target data (see Section~\ref{sec_consistency_loss}).
\fi 
%\subsubsection{Multi-Perspective Feature Enrichment Method}
%As illustrated in the bottom of Fig.~\ref{fig_pipeline}, our proposed MFE exploits target fMRI from various perspectives, \eg, window warping, receptive field manipulation, and window slicing.
Similar to the source model, each branch of MFE is equipped with a data-feeding module, a spatiotemporal feature encoder, and a class predictor. 
We initialize the MFE using parameters of the pretrained source model to facilitate source-to-target knowledge transfer without accessing source data.

\subsubsection{Feature Enrichment Strategy}

The data-feeding module of each branch corresponds to one feature enrichment perspective, and the feature encoder and class predictor share similar network architecture as the source model. 
The three feature enrichment strategies used for fMRI representation learning are introduced as follows.

(1) \textbf{\emph{Window Warping}}. 
% The value of each time point denotes BOLD signal intensity at each fMRI volume.
% In fMRI acquisition, repetition time (TR) means the time between two consecutive fMRI volumes, and it represents the temporal resolution or frequency of fMRI timeseries.
The window warping is an augmentation strategy to stretch or contract a timeseries~\cite{le2016dataaug}.
In our work, given an fMRI timeseries, we perform window warping on it with a resampling ratio $\alpha$ within the domain of $[\frac{1}{1.3}, \frac{1}{1.1}, 1, 1.1, 1.3]$,
% Here, $\alpha$ denotes the proportion based on which the time intervals are changed during the resampling process.
where $\alpha>1$ corresponds to an upsampling operation that increases the temporal resolution of fMRI timeseries, 
$\alpha<1$ corresponds to a downsampling operation that decreases the temporal resolution of fMRI timeseries, 
and $\alpha=1$ denotes that the original timeseries is retained.
All resampling operations for fMRI data are performed based on fast Fourier transform (FFT) algorithm~\cite{brigham1988fast}.
During the target model training process, we randomly select different $\alpha$ and dynamically generate resampled fMRI timeseries.
%Similar to Section~\ref{sec_datafeed_comp}, 
We also leverage an overlapping sliding window strategy (with length $l$ and stride $s$) to partition the resampled timeseries, and the derived windowed timeseries serve as the input of the spatiotemporal feature encoder.

The rationales that leverage window warping are as follows.  
(a) By upsampling the fMRI timeseries, we can obtain more data points and capture more rapid changes in brain activity, which helps the feature encoder capture fine-grained fMRI patterns. 
% Taking this rich information into account, the feature encoder can capture finer-grained fMRI patterns and model detailed temporal dynamics, which may improve feature representation learning of target data.
(b) By downsampling the fMRI timeseries, we can reduce the number of data points and capture coarse-grained fMRI patterns and more general temporal trends.
This practice may filter out random noise and make the timeseries smoother, which helps improve the signal-to-noise ratio, thus making the feature encoder capture more informative fMRI patterns. 
(c) In fMRI domain adaptation, source and target data are usually collected from different scanners/protocols, which may have different temporal resolutions.
This window warping strategy can help the feature encoder capture fMRI timeseries with varying temporal resolutions, catering to different image acquisition settings in multi-site studies. %which facilitates effective cross-domain knowledge transfer.

(2) \textbf{\emph{Receptive Field Manipulation}}.  
Many fMRI studies~\cite{liu2023spatialt, luo2021shared} leverage the sliding window scheme to characterize temporal dynamics of fMRI. %, but they often use fixed window size, limiting the ability to capture brain variability from different receptive fields. 
To further model brain variability across time, we design a unique receptive field manipulation strategy to dynamically adjust the size of sliding windows. 
% from different receptive fields, we  to  
% Here, instead of fixing the sliding window size, we propose to manipulate different window sizes to examine brain activity.
Specifically, we randomly select the window size within the range of $\beta\in\{40, 60, 80, 100\}$ during training, and the derived multi-scale windowed timeseries are fed into the spatiotemporal feature encoder for fMRI feature learning.

The rationales that we dynamically manipulate window sizes are as follows. 
Using different sizes of sliding windows enables the examination of brain activity from different scales.
Specifically, using a larger window size (\eg, 100) that covers a longer timeseries span helps the feature encoder capture more contextual fMRI patterns. % while ignoring transient temporal fluctuations in brain activity. 
Using a smaller window size (\eg, 40) can help the feature encoder detect more transient brain changes and finer-grained temporal variations.
The combination of both of them may provide a more comprehensive representation of brain variability inherent in fMRI signals.

(3) \textbf{\emph{Window Slicing}}. 
Existing studies mainly use full-length fMRI timeseries for feature extraction while ignoring augmenting timeseries by generating multiple fMRI segments from a full-length one.
We dynamically slice fMRI timeseries from the full-length one based on a ratio $\gamma\in \{85\%, 90\%, 95\%, 100\%\}$, and these sliced timeseries serve as the input of the spatiotemporal feature encoder.
Here, $\gamma=L_{s}/L$, where $L_{s}$ and $L$ mean the length (\ie, time points) of sliced timeseries and full-length timeseries of target fMRI.
At each training step, the sliced fMRI timeseries change dynamically, \ie, each with a different starting point and a different length.
%The rationales that we leverage window slicing in this work are as follows.  
Most fMRI studies usually have only a limited sample size (\eg, tens), which negatively affects the capability of the learned model.
So in this work, we propose to generate more fMRI segments based on window slicing to augment input samples, encouraging the feature encoder to learn more robust fMRI patterns to improve learning performance. 
%(2) Different slices capture different portions of the fMRI timeseries, including various temporal patterns or fluctuations, which may help the feature encoder resilient to variations in fMRI data and improve its ability to handle timeseries with different lengths and patterns.
% iii) In some cases, a small fraction of the fMRI timeseries may contain noise. And using dynamic fMRI segments may avoid these signals, leading to more reliable feature encoding.

In MFE, each branch corresponds to a specific feature enrichment strategy. 
%So far, we have constructed the multi-perspective feature enrichment method (MFE), and then 
We initialize each branch using parameters of the pretrained source model to facilitate cross-domain knowledge transfer (see Fig.~\ref{fig_pipeline}).
In this way, we can transfer discriminative fMRI representations learned from source domain to target domain, which helps yield a robust target model for improved inference based on unlabeled target data. % (detailed in Section~\ref{sec_consistency_loss}). 

\subsubsection{Mutual-Consistency Constraint}\label{sec_consistency_loss}
As mentioned above, to fully exploit unlabeled target fMRI data, we develop a feature enrichment strategy to generate multi-perspective features for each input subject. 
Intuitively, these features tend to be similar, because they are generated from the same subject. 
Accordingly, we design a \textit{mutual-consistency constraint} $L_{M}$ to train the MFE for target inference. 
The main idea is to encourage fMRI features generated from various perspectives to be consistent. 
\if false
To fully exploit unlabeled target fMRI data, we design a \textit{mutual-consistency constraint} $L_{M}$ to train the MFE for target inference.
As illustrated in the bottom right of Fig.~\ref{fig_pipeline}, we explicitly perform feature alignment across different branches to reduce their feature distribution gap. 
The main idea is to encourage fMRI features generated from various perspectives to be consistent since they are for the same subject. 
\fi
Mathematically, we propose to enforce pair-wise consistency of spatiotemporal features $\tilde{H}$ and output logits $O$ generated by three branches by minimizing the following mutual-consistency constraint:
\begin{equation}\label{eq_final_loss}
L_{M}=L_{H}+L_{O},
\end{equation}
%\vspace{-1mm}
\begin{equation}\label{eq_feature_loss}
L_{H} = \frac{1}{N_T}
\sum\nolimits_{n=1}^{N_T}\sum\nolimits_{i,j=1, j\neq i}^{m}{\|{\tilde{H}}^{i}_n - {\tilde{H}}^{j}_n\|}^{2},
\end{equation}
%\vspace{-1mm}
\begin{equation}\label{eq_logit_loss}
L_{O} = \frac{1}{N_T}
\sum\nolimits_{n=1}^{N_T}\sum\nolimits_{i,j=1, j\neq i}^{m}{\|{O}^{i}_n - {O}^{j}_n\|}^{2},
\end{equation}
where 
${\tilde{H}}^{i}_n$ and ${O}^{i}_n$ denotes spatiotemporal features and output logits generated from $i$-th feature enrichment strategy for $n$-th subject, respectively.
Here, $m$ denotes the number of feature enrichment strategies ($m=3$ in this work),  
$N_T$ denotes the number of target subjects. % in the target domain. 
% \ie, window warping, receptive field manipulation, and window slicing.
With Eq.~\eqref{eq_final_loss}, three branches of MFE (initialized by the pretrained source model) can be \emph{collaboratively fine-tuned on unlabeled target data}, thus catering to the data distribution of the target domain. 
%exploit their inherent characteristics, aiming to extract informative fMRI features. 
% In this way, the parameters of MFE (initialized by the pretrained source model) can be further fine-tuned using unlabeled target samples
%to fit the data distribution of the target data.
%With the guidance of the learned target features, parameters of MFE initialized by source model will shift towards the target domain, thus improving target inference performance. 
%Based on the trained MFE, we can perform target inference. 

\subsubsection{Inference}\label{sec_inference}
During inference, given a new test subject with fMRI timeseries, we first employ the sliding window scheme for timeseries partition, and the windowed timeseries are then fed into each branch of MFE to generate prediction logits. 
The logits are then converted to predicted probabilities based on a $sigmoid$ function, and we average the probabilities of three branches to derive final predictions.

\subsection{Unsupervised Pretraining on Large-Scale Auxiliary fMRI}\label{sec_unsup_tr}

%Existing fMRI adaptation studies are with limited sample sizes in each individual domain, which limits effective feature extraction of fMRI data.
% Second, these studies mainly focus on a limited range of diseases (most focus on a single disease), which hinders the extraction of generalized fMRI representations of various diseases. 
While there may be limited fMRI data in the source domain, there exist many large-scale auxiliary fMRI databases (even without task-specific label information) that can be employed to facilitate fMRI feature extraction. 
%To address this problem, 
To this end, we propose an \emph{unsupervised pretraining strategy} by using three large-scale fMRI databases, including  ABIDE~\cite{craddock2013neuro}, REST-meta-MDD~\cite{yan2019reduced}, and ADHD-200~\cite{bellec2017neuro}. 
% They include rs-fMRI scans of ?? healthy subjects and ?? subjects with different neurological diseases, \ie, autism spectrum disorder, major depressive disorder, and attention-deficit/hyperactivity disorder. 
From the three databases, we select subjects with fMRI length (\ie, time points) exceeding 170, resulting in 3,806 auxiliary rs-fMRI scans. 
%Note that these auxiliary fMRI data were acquired from different multiste studies that use different scanners and/or scanning protocols, and even from different populations. 
%From them, we select subjects with fMRI length (time points) exceeding 170, resulting in 3,806 rs-fMRI scans.
The demographic information of these subjects is shown in Table~\ref{tab_unsuptr_demographics}, and their ID information is listed in \emph{Supplementary Materials}\footnote{The diagnostic labels of 26 subjects in Brown site in ADHD-200 are not provided. The demographic information for 48 subjects of Site-4 in REST-meta-MDD is not provided. The gender of one subject at NYU$_2$ (ID: 10044) in ADHD-200 is not given. 180 subjects in ADHD-200 have two fMRI scans.}.

\begin{table}[tp]
\setlength{\belowcaptionskip}{-2pt}
\setlength{\abovecaptionskip}{-2pt}
\setlength{\abovedisplayskip}{-2pt}
\setlength{\belowdisplayskip}{-2pt}
    \scriptsize
    \renewcommand{\arraystretch}{1}
    % \footnotesize
    \setlength\tabcolsep{2pt}
    \caption{Demographic characteristics of the studied subjects of three auxiliary fMRI databases, \ie, ABIDE~\cite{craddock2013neuro}, REST-meta-MDD~\cite{yan2019reduced}, and ADHD-200~\cite{bellec2017neuro}. 
    ASD: autism spectrum disorder; MDD: major depressive disorder; ADHD: attention-deficit/hyperactivity disorder; HC: healthy control; M/F: Male/Female; std: standard deviation.}
    \label{tab_unsuptr_demographics}
    \centering
    \begin{tabular*}{0.48\textwidth}{@{\extracolsep{\fill}} lc cc cc cc c}
        \toprule
        \multirow{2.5}{*}{~Database}&
        \multicolumn{2}{c}{ABIDE} & & \multicolumn{2}{c}{REST-meta-MDD} & & \multicolumn{2}{c}{ADHD-200}\\
        \cmidrule(lr){2-3} \cmidrule(lr){5-6} \cmidrule(lr){8-9}
        &ASD&HC&&MDD&HC&&ADHD&HC \\
        \midrule
        ~Subject \# & 351 & 370 && 1,163 & 1,004 && 285 & 379 \\
        ~Gender (M/F) & 308/43 & 295/75 && 415/748 & 425/579 && 232/52 & 196/183  \\
        ~Age (mean$\pm$std)& 16.9±8.0 & 16.6±6.8 && 36.9±14.9 & 37.0±16.0 && 12.2±3.1 & 13.2±3.5 \\
        % ~Age (mean$\pm$std)& 16.89±7.98 & 16.57±6.76 && 36.89±14.88 & 36.98±15.97 && 12.15±3.05 & 13.18±3.48 \\
        \bottomrule
    \end{tabular*}
\end{table}

% ABIDE: 721, 721 scans.
% MDD: 2215, 2215 scans.
% ADHD: 664+26 subjects (Brown: 26 subjects without diagnostic label), 870 scans.

In our unsupervised pretraining strategy, we leverage the same network architecture as MFE introduced in Section~\ref{sec_tribranch}, and use these auxiliary fMRI data as input for model pretraining in an unsupervised manner.
%In this way, we can partly alleviate the small sample size issue in fMRI adaptation. 
\if false
Our unsupervised pretraining strategy leverages the same architecture as MFE introduced in Section~\ref{sec_tribranch}, while their only difference lies in the input data. 
Specifically, the unlabeled target data serve as inputs in Section~\ref{sec_tribranch}, while in unsupervised pretraining, the inputs are 3,806 unlabeled fMRI data from three large-scale auxiliary databases.
Similarly, we also design a mutual-consistency constraint (same as $L_{M}$) to train this unsupervised pretraining model, which minimizes feature distribution gap across branches for the same subject.
\fi 
The pretrained MFE will be used to initialize the source model. 
Since these auxiliary fMRI data are acquired from multi-site studies that use different scanners and/or scanning protocols, and even from different diseases and populations, this pretraining strategy is expected to help produce a general fMRI feature encoder. 
%By pretraining our model on these large-scale auxiliary fMRI data, we can obtain a generalized fMRI feature encoder, which learns rich fMRI features that may not be captured from one single domain.
%Besides, data from multiple databases exhibit much diversity (\eg, different population demographics), which can improve the generalizability of the feature encoder.
% In this way, the fMRI features generated from this encoder are prone to be less affected by specific domain bias.
%It is expected that the 
Accordingly, features learned from this general feature encoder will be less affected by domain bias and more transferable to downstream tasks, helping address cross-domain shift problem. 
%, helping reducing its search space. 
Our pretrained model has been released to the public\footnote{\url{https://github.com/yqfang9199/SCDA}}.

%\subsection{Implementation}\label{sec_implementation} 
%We first construct an initial MFE by performing unsupervised pretraining on 3,806 unlabeled auxiliary fMRI. 
%Then we average parameters of three branches of MFE to initialize a source model. %, which is trained with labeled source data in a supervised manner.
%After obtaining the pretrained source model, we use its parameters to reinitialize feature encoders and class predictors of MFE, and optimize MFE using unlabeled target data via $L_M$. 
 %, while the hyperparameters for feature enrichment (\ie, $\alpha$, $\beta$, and $\gamma$) are set as $\alpha$=[], $\beta$, and $\gamma$.
\if false 
During inference, given a new test subject with fMRI timeseries, we first employ the sliding window scheme for timeseries partition, and the windowed timeseries are then fed into each branch of MFE to generate prediction logits.
The logits are then converted to predicted probabilities based on a $sigmoid$ function, and we average the probabilities of three branches to derive final prediction.
\fi
% Note that each branch of the framework can be used for inference, and in practice we find ...
% is used for prediction in the experiments if not specified.
The proposed SCDA is implemented via PyTorch with Adam optimizer (mini-batch size: 64).
The initial learning rate is set to 0.0003 and dropped by 0.5 every 50 epochs, with a total of 150 training epochs. 
%Several key hyperparameters are set as $l=40$, $s=30$, $D=64$.
The experiments are performed on a GPU (NVIDIA TITAN Xp) with 12GB of memory. 
%The source code of SCDA will be available upon acceptance.

\begin{table*}[tp]
\setlength{\belowcaptionskip}{-2pt}
\setlength{\abovecaptionskip}{-2pt}
\setlength{\abovedisplayskip}{-2pt}
\setlength{\belowdisplayskip}{-2pt}
%    \footnotesize
\scriptsize
    \setlength\tabcolsep{2pt}
    \caption{Demographic information of the studied subjects of two groups of cross-scanner prediction tasks:
    1) Two largest sites (\ie, NYU$_1$ and UM) of ABIDE~\cite{craddock2013neuro} are used to identify ASD patients from HCs, where NYU$_1$ serves as the source domain and UM is used as the target domain.
    2) Two largest sites (\ie, NYU$_2$ and Peking) of ADHD-200~\cite{bellec2017neuro} are used to identify ADHD patients from HCs, where NYU$_2$ and Peking are regarded as source and target domains, respectively.
    ASD: autism spectrum disorder; 
    ADHD: attention-deficit/hyperactivity disorder; 
    HC: healthy control; 
    M/F: Male/Female;
    std: standard deviation.
    }
    \label{tab_demographic_cross_scanner}
    \centering
    \resizebox{1\textwidth}{!}{
    \begin{tabular*}{1.0\textwidth}{@{\extracolsep{\fill}} l|ccccccccccc}
    \toprule
    \multirow{4}{*}{~Group}&    
    \multicolumn{5}{c}{Cross-scanner Prediction with ABIDE} & \phantom{ab} && \multicolumn{3}{c}{Cross-scanner Prediction with ADHD-200} & \phantom{ab} \\    
    \cmidrule(lr){2-6} \cmidrule(lr){8-12}
    & \multicolumn{2}{c}{NYU$_1$ (source domain)} & \phantom{ab} & \multicolumn{2}{c}{UM (target domain)} & \phantom{ab} & \multicolumn{2}{c}{NYU$_2$ (source domain)} & \phantom{ab} & \multicolumn{2}{c}{Peking (target domain)} \\% & \phantom{ab} \\
    \cmidrule(lr){2-3} \cmidrule(lr){5-6} \cmidrule(lr){8-9} \cmidrule(lr){11-12}
    & ASD & HC && ASD & HC && ADHD & HC && ADHD & HC \\
    \midrule
    ~Subject No. & 75 & 100 && 66 & 74 && 152 & 111 && 102 & 143 \\
    ~Gender (M/F) & 65/10 & 74/26 && 57/9 & 56/18 && 115/36 & 55/56 && 90/12 & 84/59 \\
    ~Age (mean±std)& 14.74±7.08 & 15.65±6.16 && 13.17±2.40 & 14.81±3.61 && 10.97±2.67 & 12.11±3.10 && 12.09±2.04 & 11.43±1.86  \\
    ~Time points ($L$) & 176 & 176 && 296 & 296 && 171 & 171 && 231 & 231  \\
    \bottomrule
    \end{tabular*}}
\end{table*}

%-------------------------------------------------------------------

% \ie, ABIDE~\cite{craddock2013neuro}, REST-meta-MDD~\cite{yan2019reduced}, and ADHD-200~\cite{bellec2017neuro}

\begin{table*}[tp]
\setlength{\belowcaptionskip}{-2pt}
\setlength{\abovecaptionskip}{-2pt}
\setlength{\abovedisplayskip}{-2pt}
\setlength{\belowdisplayskip}{-2pt}
\scriptsize
    \setlength\tabcolsep{2pt}
    \caption{Demographic information of the studied subjects of two groups of cross-study prediction tasks:
    1) The largest site (\ie, Site-21) of REST-meta-MDD~\cite{yan2019reduced} is used as source domain and the largest site (\ie, NYU$_1$) of ABIDE~\cite{craddock2013neuro} serves as target domain.
    2) The largest site (\ie, NYU$_2$) of ADHD-200~\cite{bellec2017neuro} is regarded as the source domain and T2DM is used as the target domain.
    MDD: major depressive disorder;
    ASD: autism spectrum disorder; 
    ADHD: attention-deficit/hyperactivity disorder; 
    T2DM: Type 2 diabetes mellitus;
    T2DM-CI: T2DM with cognitive impairment; 
    HC: healthy control; 
    M/F: Male/Female;
    std: standard deviation.
    }
    \label{tab_demographic_cross_disease}
    \centering
    \resizebox{1\textwidth}{!}{
    \begin{tabular*}{1.0\textwidth}{@{\extracolsep{\fill}} l|ccccccccccc}
    \toprule
    \multirow{4}{*}{~Group}&
    \multicolumn{5}{c}{Cross-study Prediction with REST-meta-MDD and ABIDE} & \phantom{ab} && \multicolumn{3}{c}{Cross-study Prediction with ADHD-200 and T2DM} \\
    \cmidrule(lr){2-6} \cmidrule(lr){8-12}
& \multicolumn{2}{c}{Site-21 (source domain)} & \phantom{ab} & \multicolumn{2}{c}{NYU$_1$ (target domain)} & \phantom{ab} & \multicolumn{2}{c}{NYU$_2$ (source domain)} & \phantom{ab} & \multicolumn{2}{c}{T2DM (target domain)} \\%& \phantom{ab} \\
    \cmidrule(lr){2-3} \cmidrule(lr){5-6} \cmidrule(lr){8-9} \cmidrule(lr){11-12}
    & MDD & HC && ASD & HC && ADHD & HC && T2DM-CI & HC \\
    \midrule
    ~Subject No. & 282 & 251 && 75 & 100 && 152 & 111 && 30 & 31  \\
    ~Gender (M/F) & 99/183 & 87/164 && 65/10 & 74/26 && 115/36 & 55/56 && 20/10 & 21/10  \\
    ~Age (mean±std)&38.74±13.74&39.64±15.87&& 14.74±7.08 & 15.65±6.16 && 10.97±2.67 & 12.11±3.10 && 52.15±8.29 & 50.48±7.07 \\
    ~Time points ($L$) & 232 & 232 && 176 & 176 && 171 & 171 && 950 & 950  \\
    \bottomrule
    \end{tabular*}}
\end{table*}

\section{Experiments}\label{sec_experiment}
\subsection{Materials}
\subsubsection{Data Acquisition and Preprocessing}

Three public datasets and one private dataset are employed in this work.  
1) \textbf{\emph{ABIDE}}~\cite{craddock2013neuro}: This dataset contains 1,112 rs-fMRI scans acquired from 16 international imaging sites, including 539 subjects diagnosed with autism spectrum disorder (ASD) and 573 healthy controls (HCs).
2) \textbf{\emph{REST-meta-MDD}}~\cite{yan2019reduced}: This database contains 2,428 rs-fMRI data acquired from 25 research groups, including 1,300 subjects suffering from major depressive disorder (MDD) and 1,128 HCs.
3) \textbf{\emph{ADHD-200}}~\cite{bellec2017neuro}: This dataset is a collaboration of 8 imaging sites with 1,395 rs-fMRI scans acquired from 388 subjects diagnosed with attention-deficit/hyperactivity disorder (ADHD) and 585 HCs. 
4) \textbf{\emph{T2DM}}: This rs-fMRI dataset is collected from The First Affiliated Hospital of Guangzhou University of Chinese Medicine, which
aims to investigate Type 2 diabetes mellitus-associated cognitive impairment.
Specifically, a Siemens (Munich, Germany) 3T Prisma scanner with a 64-channel head coil is used to acquire fMRI scans.
The gradient echo-planar imaging sequence acquisition parameters are as follows: repetition time (TR) is $500\,{ms}$;
echo time (TE) is $30\,{ms}$; field-of-view is $244\times244$; slice thickness is $3.5\,{mm}$; voxel size is $3.5\times3.5\times3.5\,{mm}^3$.
T2DM consists of 30 subjects with cognitive impairment (T2DM-CI) and 31 HCs.

A total of 3,806 auxiliary rs-fMRI scans from three datasets (\ie, ABIDE, REST-meta-MDD, and ADHD-200) are used for unsupervised pretraining, where no label information of these data is used. 
\if false
The large-scale auxiliary fMRI mentioned in Section~\ref{sec_unsup_tr} are from three public fMRI databases introduced above, \ie, ABIDE, REST-meta-MDD, and ADHD-200.
It is noted that the label information of these auxiliary fMRI data is unavailable.
\fi 
These auxiliary fMRIs are used to construct an initial MFE to obtain a general encoder, and the learned network parameters are used to initialize the source model. 
%The demographic information of these subjects is shown in Table~\ref{tab_unsuptr_demographics}, and the subject ID information is listed in \emph{Supplementary Materials}.  

The rs-fMRIs are preprocessed using a standard pipeline, with several key steps listed as follows:  
1) magnetization equilibrium, %discarding the first several volumes, 
2) slice timing correction, 
3) head motion correction, 
4) bandpass filtering, % ($0.01$-$0.10\,Hz$), 
5) nuisance signal removal, 
6) registration between T1-weighted MRI and mean functional images, 
7) spatial normalization to Montreal Neurological Institute (MNI) template, and 
8) brain partition into $N$=116 ROIs based on automated anatomical atlas (AAL).
We use regional mean timeseries to represent each subject.
% For each ROI of each subject, we calculate its mean time series by averaging the BOLD signals over all voxels within this ROI.

\subsubsection{Domain Setting}
%In the experiments, w
We perform \textbf{\emph{cross-scanner}} and \textbf{\emph{cross-study}} prediction to validate the efficacy of SCDA in cross-domain fMRI adaptation, where \emph{domain} denotes \emph{scanner} and \emph{study}, respectively. 
The demographic characteristics of the studied subjects in cross-scanner and cross-study prediction are given in Table~\ref{tab_demographic_cross_scanner} and Table~\ref{tab_demographic_cross_disease}, respectively. 

1) \textbf{Cross-scanner prediction}. 
In multi-site studies, different hospitals/sites may use different scanners and imaging protocols to acquire fMRI data, leading to data distribution differences across sites.
These differences may negatively affect the performance and generalizability of models learned from one site when applied to another site.
Cross-scanner prediction is essential to reduce distribution shifts between sites. % and thus make the learned model more robust and transferable.
Here, we conduct two groups of experiments to investigate cross-scanner prediction: 
1) Two largest sites (\ie, NYU$_1$ and UM) of ABIDE are used to identify ASD patients from HCs, where NYU$_1$ with 175 subjects serves as the source domain and UM with 140 subjects is used as the target domain.
Data of NYU$_1$ and UM are collected by using a Siemens Magnetom Allegra MRI scanner and a GE Signa MRI scanner, respectively.
% \url{https://fcon_1000.projects.nitrc.org/indi/abide/abide_I.html}
2) Two largest sites (\ie, NYU$_2$ and Peking) of ADHD-200 are used to identify ADHD patients from HCs, where NYU$_2$ with 263 subjects is employed as the source domain and Peking with 245 subjects serves as the target domain.
Data of NYU$_2$ and Peking are collected by using a Siemens Magnetom Allegra MRI scanner and a Siemens Magnetom Trio Tim MRI scanner, respectively.
% \url{http://fcon_1000.projects.nitrc.org/indi/adhd200/#}

2) \textbf{Cross-study prediction}.
Many studies have limited sample sizes due to various reasons such as disease rarity and patient dropout, while some other studies may have relatively large data.  %limited patient accessibility. %, etc.
Cross-study prediction allows for leveraging labeled data from one study to assist in learning models for another study with limited sample sizes.
Here, we conduct two groups of experiments to investigate cross-study prediction:  
% The \textbf{\emph{cross-study}} setting also includes two groups of experiments, where source and target domains involve different brain neurological diseases:
1) The largest site (\ie, Site-21, with 533 subjects) of REST-meta-MDD is used as the source domain and the largest site (\ie, NYU$_1$, with 175 subjects) of ABIDE serves as the target domain.
2) The largest site (\ie, NYU$_2$, with 263 subjects) of ADHD-200 is used as the source domain and the T2DM (with 61 subjects) is employed as the target domain. 
%Note that in both experiments, besides the disease difference, the MRI scanners used in source and target domains are also different, which poses more challenges in target model construction.
% NYU2: Siemens MAGNETOM Allegra syngo MR 2004A
% T2DM: Siemens (Munich, Germany) 3T Prisma scanner
% MDD: a Siemens Tim Trio 3T MRI scanne
% NYU1: a Siemens Magnetom Allegra MRI scanner
%The demographic characteristics of the studied subjects in cross-scanner and cross-study prediction are given in Table~\ref{tab_demographic_cross_scanner} and Table~\ref{tab_demographic_cross_disease}, respectively.
% (TP#, scanner, number, gender, age, edu)

\subsection{Competing Methods}\label{sec_competing_methods}

We compare our proposed SCDA with five traditional methods, including 1) NF-SVM, 2) GF-SVM, 3) PCC-XGB, 4) PCC-RF, 
5) MaLRR~\cite{wang2019identifying}, and six deep learning methods, including 
6) DANN~\cite{ganin2016domain}, 
7) RAINCOAT~\cite{he2023domainseries},
8) UFA-Net~\cite{fang2023unsupervised}, 
9) SHOT~\cite{liang2020we}, 
10) CDCL~\cite{wang2022crossda}, 
and 11) SCDA-Naive,
with details introduced as follows.

1) \textbf{NF-SVM}: In this method, we first construct a functional connectivity (FC) matrix for each subject based on PCC which measures linear correlation between paired ROIs.
Then, we extract six node-based features (\ie, clustering coefficient, node strength, local efficiency, eigenvector centrality, modularity, and node betweenness centrality) from the FC matrix and perform prediction using a support vector machine (SVM).

2) \textbf{GF-SVM}: Based on the same FC matrix in NF-SVM, this method extracts five graph-based features (\ie, density, global efficiency, assortativity coefficient, characteristic path length, and transitivity) and performs prediction using an SVM. 

3) \textbf{PCC-XGB}: In this method, we first build an FC matrix based on PCC, and then we flatten the upper triangle elements of the FC matrix and convert them into a one-dimensional vectorized representation.
An XGBoost model~\cite{chen2016xgboost} is then used to produce prediction results.

4) \textbf{PCC-RF}: This method uses the same features as PCC-XGB but with a random forest classifier for final prediction.

5) \textbf{MaLRR}: It tackles multi-site domain adaptation based on rs-fMRI signals via low-rank representation (LRR) decomposition. 
Specifically, it first captures spatiotemporal fMRI features of source and target domains and transforms them into a common feature space via LRR, and then each source sample is linearly represented using all target samples.
The transformed source and target features are then fed into an SVM model for final classification.

6) \textbf{DANN}: This is a widely used unsupervised domain adaptation method.
For a fair comparison, we replace the original feature encoder with our spatiotemporal feature encoder and randomly capture fMRI features from various perspectives (same as our SCDA).
A domain discriminator is then used to differentiate which features come from which domain, thereby encouraging domain invariance in feature space.

7) \textbf{RAINCOAT}: This recently proposed method is designed to tackle domain adaptation for timeseries data.
Specifically, RAINCOAT takes advantage of both time and frequency characteristics during feature encoding in order to exploit time-frequency features of timeseries data.
With these encoded features, a Sinkhorn divergence is then leveraged to perform feature alignment between source and target domains.

8) \textbf{UFA-Net}: This method is specifically designed to address unsupervised domain adaptation for rs-fMRI data.
Specifically, it employs an attention-based graph convolutional network to capture fMRI representations for both source and target domains, and then a discrepancy-constrained module is introduced to align cross-domain feature representations.

9) \textbf{SHOT}: This method performs source-free domain adaptation only using a pretrained source model and unlabeled target data (same as our SCDA).
Specifically, SHOT leverages knowledge stored in the source model to exploit target-specific representations via both information maximization and clustering-based pseudo-labeling. 
%As our SCDA does, we also leverage unsupervised pretraining based on the three auxiliary fMRI databases to initialize the source model in SHOT.
Similar to SCDA, this method leverages unsupervised pretraining based on the three auxiliary fMRI databases to initialize the source model.

10) \textbf{CDCL}: This method also proposes a source-free domain adaptation strategy.
Specifically, CDCL first generates source prototypical features from a pretrained source model to represent cluster centers of different categories.
Then, a contrastive self-supervised learning framework is designed for cross-domain feature alignment, which aims to pull features of similar categories closer and push features of different categories away in feature space. 
Similarly, we also utilize unsupervised pretraining based on three auxiliary fMRI databases for source model initialization.

11) \textbf{SCDA-Naive}: It represents the SCDA
without unsupervised pretraining, \ie, the source
model is trained from scratch rather than initialized from large-scale fMRI databases.

Note that the four traditional methods (\ie, NF-SVM, GF-SVM, PCC-XGB, PCC-RF) only access source data for model training, while the learned models are directly applied to the target domain during inference.
The four unsupervised domain adaptation methods (\ie, MaLRR, DANN, RAINCOAT, UFA-Net) need to access labeled source data and unlabeled target data for target model construction.
%Similar to our SCDA, 
The four source-free unsupervised domain adaptation methods (\ie, SHOT, CDCL, SCDA-Naive, and SCDA) only use a pretrained source model and unlabeled target data for target model learning. 
All deep learning methods are repeated five times independently to reduce the bias caused by parameter initialization.

\begin{table*}[!t]
\setlength{\belowcaptionskip}{-2pt}
\setlength{\abovecaptionskip}{-2pt}
\setlength{\abovedisplayskip}{-2pt}
\setlength{\belowdisplayskip}{-2pt}
\footnotesize
\textbf{\renewcommand{\arraystretch}{1}}
	\caption{Classification results of twelve methods in two groups of cross-scanner prediction tasks.
	$*$ denotes that the difference between our SCDA and the competing method is statistically significant (with $p$-value$<$0.05). Best results are shown in bold.}
	\label{tab_competing_methods_cross_scanner}
	\centering
	\setlength{\tabcolsep}{0.5pt}
        \resizebox{\textwidth}{!}{
        \begin{tabular*}{1.2\textwidth}{@{\extracolsep{\fill}}l|cccccc c cccccc}\toprule
	\multirow{2}{*}{Method} & \multicolumn{6}{c}{Task 1: ASD vs. HC classification on UM} && \multicolumn{6}{c}{Task 2:  ADHD vs. HC classification on Peking} \\
        \cmidrule(l){2-7} \cmidrule(l){9-14}
		 & AUC (\%) & ACC (\%) & F1 (\%) & SEN (\%) & SPE (\%) & PRE (\%) && AUC (\%) & ACC (\%) & F1 (\%) & SEN (\%) & SPE (\%) & PRE (\%) \\
	\midrule
        NF-SVM & 53.83$^*$ & 51.43 & 54.05 & 60.61 & 43.24 & 48.78 && 55.38$^*$ & 55.92 & 47.06 & 47.06 & 62.24 & 47.06 \\ 
        GF-SVM & 50.57$^*$ & 50.71 & 54.30 & 62.12 & 40.54 & 48.24 && 52.87$^*$ & 53.88 & 44.88 & 45.10 & 60.14 & 44.66 \\ 
        PCC-XGB & 57.76$^*$ & 53.57 & 55.78 & 62.12 & 45.95 & 50.62 && 55.74$^*$ & 54.29 & 49.55 & 53.92 & 54.55 & 45.83 \\ 
        PCC-RF & 64.99$^*$ & 61.43 & 57.81 & 56.06 & 66.22 & 59.68 && 54.37$^*$ & 54.29 & 42.86 & 41.18 & 63.64 & 44.68 \\
        MaLRR & 66.57$^*$ &	62.14 &	61.31 &	63.64 &	60.81 &	59.15 && 58.82$^*$ & 57.55 & 49.02 & 49.02 & \textbf{63.64} & 49.02 \\
        DANN & 67.22±3.04$^*$ & 62.57±2.46 & 61.98±5.93 & 67.58±16.78 & 58.11±19.11 & 62.33±10.36 && 59.70±2.34$^*$ & 53.55±6.67 & 43.35±16.87 & 53.92±33.77 & 53.29±34.23 & 48.16±4.75 \\
        RAINCOAT & 65.11±3.37$^*$ & 58.71±3.21 & 56.08±11.02 & 60.61±20.53 & 57.03±19.37 & 57.07±4.41 && 54.92±2.42$^*$ & 53.88±2.18 & 48.85±2.53 & 53.14±5.66 & 54.41±6.48 & 45.50±1.82 \\
        UFA-Net & 54.36±2.10$^*$ & 54.71±1.84 & 65.95±2.33 & 52.44±2.49 & 56.10±1.48 & 34.05±2.33 && 53.24±1.66$^*$ & 52.24±2.93 & 47.05±8.12 & 53.73±16.48 & 51.19±15.70 & 44.01±1.74 \\
        SHOT & 67.84±3.90$^*$ & 55.57±5.39 & 53.99±18.92 & 67.58±31.62 & 44.86±33.43 & 57.35±9.76 && 56.23±0.77$^*$ & 54.12±1.64 & 50.45±0.96 & 56.08±0.96 & 52.73±2.95 & 45.88±1.48 \\
        CDCL & 68.10±1.13$^*$ & 63.00±1.88 & 61.67±4.69 & 64.85±13.53 & 61.35±14.34 & 61.20±4.75 && 58.22±1.53$^*$ & 53.63±6.29 & 45.15±16.03 & 55.88±31.48 & 52.03±32.12 & 47.55±6.98 \\        
        SCDA-Naive & 66.70±4.06$^*$ & 62.71±5.26 & 63.34±4.60 & 68.48±8.10 & 57.57±11.10 & 59.63±6.05 && 58.04±2.17$^*$ & 53.88±5.97 & 45.95±13.02 & 54.90±28.70 & 53.15±29.97 & 48.86±7.10 \\
        SCDA (Ours) & \textbf{72.79±2.10} & \textbf{68.00±0.83} & \textbf{66.77±2.74} & \textbf{68.79±7.94} & \textbf{67.30±6.53} & \textbf{65.48±1.92} && \textbf{61.56±0.93} & \textbf{58.20±3.78} & \textbf{51.18±7.32} & \textbf{56.27±20.16} & 59.58±20.79 & \textbf{52.23±5.79} \\
	\bottomrule
	\end{tabular*}}
\end{table*}

\subsection{Experimental Settings}\label{sec_experimental_settings}
Six evaluation metrics are used to validate the effectiveness of our SCDA, including the area under the receiver operating characteristic curve (AUC), classification accuracy (ACC), F1 score (F1), sensitivity (SEN), specificity (SPE), and precision (PRE). 
Denote true positive $(TP)$ as the number of subjects that are correctly classified as the positive category (\ie, disease group), true negative $(TN)$ as the number of subjects that are correctly classified as the negative category (\ie, HC), false positive $(FP)$ as the number of subjects that are wrongly classified as the positive group, and false negative $(FN)$ as the number of subjects that are wrongly classified as the negative category.
Here, ACC = $\frac{TP+TN}{TP + FP + TN + FN}$, F1 = $\frac{2TP}{2TP + FP + FN}$, SEN = $\frac{TP}{TP+FN}$, SPE = $\frac{TN}{TN+FP}$, and PRE =$ \frac{TP}{TP+FP}$. 

\subsection{Results of Cross-Scanner Prediction}

\begin{table*}[!t]
\setlength{\belowcaptionskip}{-2pt}
\setlength{\abovecaptionskip}{-2pt}
\setlength{\abovedisplayskip}{-2pt}
\setlength{\belowdisplayskip}{-2pt}
\footnotesize
\textbf{\renewcommand{\arraystretch}{1}}
	\caption{Classification results of twelve methods in two groups of cross-study prediction tasks.
	$*$ denotes that the difference between our SCDA and the competing method is statistically significant (with $p$-value$<$0.05). Best results are shown in bold.}
	\label{tab_competing_methods_cross_disease}
	\centering
	\setlength{\tabcolsep}{0.5pt}
        \resizebox{\textwidth}{!}{
        \begin{tabular*}{1.2\textwidth}{@{\extracolsep{\fill}}l|cccccc c cccccc}\toprule
	\multirow{2}{*}{Method} & \multicolumn{6}{c}{Task 1: ASD vs. HC classification on NYU$_1$} && \multicolumn{6}{c}{Task 2: T2DM-CI vs. HC on T2DM} \\
        \cmidrule(l){2-7} \cmidrule(l){9-14}
		 & AUC (\%) & ACC (\%) & F1 (\%) & SEN (\%) & SPE (\%) & PRE (\%) && AUC (\%) & ACC (\%) & F1 (\%) & SEN (\%) & SPE (\%) & PRE (\%) \\
	\midrule
        NF-SVM & 51.04$^*$ & 54.86 & 47.68 & 48.00 & 60.00 & 47.37 && 55.16$^*$  & 54.10  & 57.58  & 63.33  & 45.16  & 52.78 \\
        GF-SVM & 51.67$^*$ & 55.43 & 56.18 & 66.67 & 47.00 & 48.54 && 54.52$^*$  & 52.46  & 56.72  & 63.33  & 41.94  & 51.35 \\
        PCC-XGB & 53.20$^*$ & 49.71 & 49.43 & 57.33 & 44.00 & 43.43 && 54.73$^*$  & 55.74  & 57.14  & 60.00  & 51.61  & 54.55 \\
        PCC-RF & 54.43$^*$ & 54.29 & 47.37 & 48.00 & 59.00 & 46.75 && 54.30$^*$  & 55.74  & 55.74  & 56.67  & 54.84  & 54.84 \\
        MaLRR & 71.69$^*$ & 66.86 & 63.75 & 68.00 & 66.00 & 60.00 && 55.11$^*$  & 52.46  & 56.72  & 63.33  & 41.94  & 51.35 \\
        DANN & 63.92±3.00$^*$ & 59.20±4.94 & 56.21±5.36 & 63.20±17.25 & 56.20±20.39 & 53.85±6.04 && 59.70±3.67$^*$ & 54.43±4.32 & 47.10±16.77 & 50.00±30.98 & 58.71±31.30 & 55.26±6.12 \\
        RAINCOAT & 54.60±3.90$^*$ & 51.77±3.04 & 50.02±4.36 & 57.60±12.38 & 47.40±13.92 & 45.53±2.44 && 54.67±5.58$^*$ & 55.08±3.68 & 57.35±8.94 & 64.67±18.45 & 45.81±13.75 & 53.46±2.74 \\
        UFA-Net & 56.18±3.36$^*$ & 55.09±1.06 & 62.80±4.26 & 47.48±1.27 & 60.29±0.80 & 37.20±4.26 && 56.17±2.55$^*$ & 52.79±3.34 & 34.19±6.64 & 51.36±2.34 & 56.27±6.21 & 65.81±6.64 \\
        SHOT & 72.78±0.50$^*$ & 65.49±2.99 & 63.76±1.82 & \textbf{71.20±8.71} & 61.20±11.07 & 58.71±4.37 && 75.38±0.98 & 62.30±4.28 & 55.60±22.62 & 62.67±34.15 & 61.94±27.78 & \textbf{71.34±16.52} \\
        CDCL & 72.89±0.66$^*$ & 65.60±2.62 & 61.58±5.41 & 66.93±17.23 & 64.60±16.70 & 60.95±6.81 && 74.75±1.62$^*$ & 64.59±1.67 & 63.31±9.26 & 66.67±19.66 & 62.58±16.27 & 65.28±6.06 \\        
        SCDA-Naive & 65.48±2.02$^*$ & 59.09±4.19 & 56.62±6.27 & 65.60±19.93 & 54.20±21.58 & 54.07±6.38 && 62.06±4.19$^*$ & 56.07±5.81 & 52.83±8.04 & 53.33±19.66 & 58.71±27.93 & 60.33±11.41 \\
        SCDA (Ours) & \textbf{73.31±0.53} & \textbf{67.20±2.25} & \textbf{63.85±1.48} & 68.00±8.43 & \textbf{66.60±10.13} & \textbf{61.28±4.43} && \textbf{75.63±0.66} & \textbf{65.57±2.74} & \textbf{65.74±2.60} & \textbf{68.00±10.87} & \textbf{63.23±14.80} & 66.16±8.75 \\
	\bottomrule
	\end{tabular*}}
\end{table*}

The classification results of our SCDA and eleven competing methods in cross-scanner prediction tasks are reported in Table~\ref{tab_competing_methods_cross_scanner}.
We also perform a two-tailed paired $t$-test on the results of SCDA and each competing method, and `*' denotes that the performance of SCDA is statistically significantly different ($p<0.05$) from a specific method. 
From Table~\ref{tab_competing_methods_cross_scanner}, we have the following interesting observations.

\emph{First}, our SCDA outperforms the first four traditional models (\ie, NF-SVM, GF-SVM, PCC-XGB, PCC-RF) by a large margin.
The possible reason is that these methods only consider handcrafted features derived from a stationary functional connectivity matrix, without taking advantage of spatiotemporal characteristics inherent in fMRI timeseries.
Despite the MaLRR method can %(\ie, the fifth traditional method) 
encode spatiotemporal features, its feature learning and class prediction are performed in a separate manner, which may degrade the learning performance. 
\emph{Second}, the SCDA generally achieves better results compared with three deep learning-based domain adaptation approaches (\ie, RAINCOAT, UFA-Net, and DANN).
Note that RAINCOAT is originally designed for timeseries data and UFA-Net is specifically designed for fMRI data.
Compared with these two methods, our SCDA exploits fMRI timeseries from more perspectives, resulting in general and diverse fMRI representations, which may facilitate effective cross-domain knowledge transfer.
\emph{Furthermore}, the two source-free domain adaptation approaches (\ie, SHOT and CDCL) show inferior performance compared with our SCDA. 
% For instance, compared with SHOT, the SCDA improves the ACC and AUC values by xx\% and xx\% in ASD vs. HC classification task, respectively.
%Specifically, the possible reason why SHOT does not yield promising results may be that SHOT is optimized with information maximization, which may assign a high confidence score to incorrect prediction, leading to suboptimal results. 
The possible reason may be that SHOT is optimized with information maximization, which may assign a high confidence score to incorrect prediction, leading to suboptimal results.
And the CDCL method generates source prototypical features from a pretrained source model to initialize category centers of target data, which may introduce bias when there exists a large domain gap.
This bias could negatively affect the adaptation process, as the initial centers may not accurately represent the target domain.
Compared with these two methods, our SCDA neither utilizes information maximization nor category center initialization. 
Instead, the SCDA introduces a multi-perspective feature enrichment method to exploit fMRI characteristics inherent in target data based on stored source knowledge, demonstrating its superior performance.
\emph{Additionally}, our SCDA shows better results compared with SCDA-Naive which does not use unsupervised pretraining.
The reason may be that, with unsupervised pretraining, SCDA can learn a general fMRI feature encoder from large-scale and diverse fMRI databases.
And features derived from this encoder are expected to be less affected by domain bias, helping address cross-domain shift issues and improve target inference.

\begin{figure*}[!t]\centering
\setlength{\belowcaptionskip}{-2pt}
\setlength{\abovecaptionskip}{-2pt}
\setlength{\abovedisplayskip}{-2pt}
\setlength{\belowdisplayskip}{-2pt}
	\includegraphics[width=0.98\textwidth]{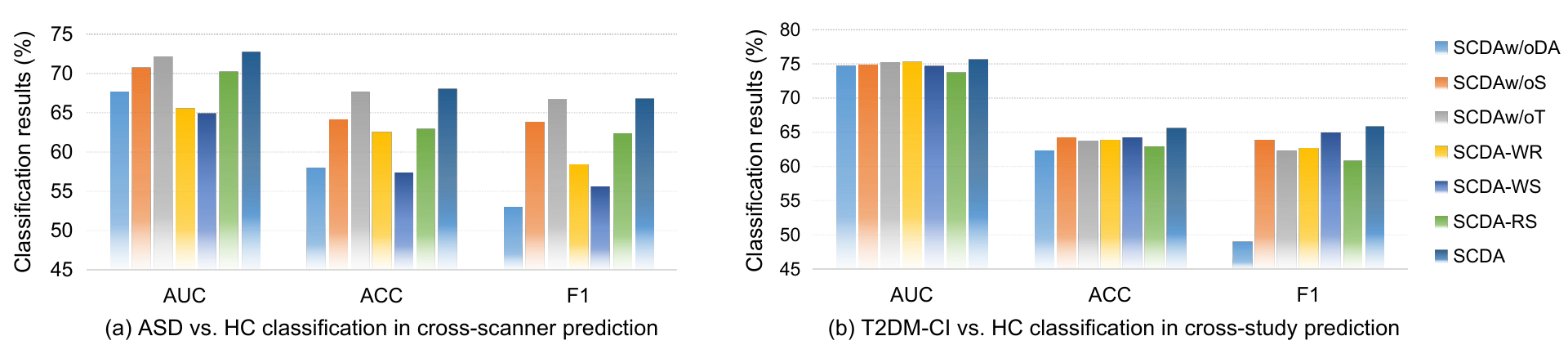}
	\caption{Experimental results (AUC, ACC, and F1) of the proposed SCDA and its six variants in (a) identifying ASD patients from HCs in UM in cross-scanner prediction task, and (b) classifying T2DM-CI patients from HCs in T2DM in cross-study prediction task.}\label{fig_ablation_study}
\end{figure*}

\subsection{Results of Cross-Study Prediction}

The classification results of our SCDA and eleven competing methods in cross-study prediction tasks are shown in Table~\ref{tab_competing_methods_cross_disease}.
From Table~\ref{tab_competing_methods_cross_disease}, we have a similar observation to those in Table~\ref{tab_competing_methods_cross_scanner}, that is, the proposed SCDA produces the overall best performance in most cases. 
\emph{Besides}, we find that the standard deviation of our SCDA is smaller than that of the most competing methods, which indicates that the SCDA yields more stable and consistent classification results under different parameter initialization, suggesting the robustness of our method.
\emph{In addition}, we observe that SCDA-Naive has comparative results with DANN which has access to labeled source domain and unlabeled target domain.
Note that in our work, we try to make SCDA-Naive and DANN as comparable as possible.
Specifically, both of them extract fMRI features from multiple perspectives (\ie, window warping, receptive field manipulation, and window slicing), and they are both trained from scratch without the unsupervised pretraining strategy.
Under such circumstances, our SCDA-Naive trained only based on a pretrained source model and unlabeled target data can achieve competitive results with DANN, which further demonstrates its effectiveness.
\emph{Futhermore}, our SCDA generally produces promising classification performance in cross-study prediction tasks, where the target domain is usually with a limited sample size (\eg, 61 in T2DM).
This shows that our method can effectively improve inference of a small target dataset only based on a pretrained source model, which helps preserve privacy concerns and save storage burdens.

\section{Discussion}\label{sec_discussion}
\subsection{Ablation Study}

We conduct ablation studies to investigate the influence of several key components of the proposed SCDA, by comparing SCDA with its six variants called {SCDAw/oDA}, {SCDAw/oS}, {SCDAw/oT}, {SCDA-WR}, {SCDA-WS}, and {SCDA-RS}. %which are detailed as follows.
Specifically, \textbf{SCDAw/oDA} denotes the SCDA without any adaptation technique, that is, the model trained on the source domain is directly applied to the target domain. 
% \textbf{SCDAw/oU} represents the SCDA without the unsupervised pretraining strategy, that is, the source model is trained from scratch rather than initialized from large-scale fMRI databases.
\textbf{SCDAw/oS} means SCDA without spatial attention across ROIs, and \textbf{SCDAw/oT} means SCDA without temporal attention across different time windows.
% The \textbf{SCDA-O} represents that only original full-length fMRI timeseries are used as inputs of three branches in multi-perspective feature enrichment framework.
% The \textbf{SCDA-L}, \textbf{SCDA-T}, and \textbf{SCDA-W} denote that SCDA captures fMRI representations from only one perspective, \eg, window warping (W), receptive field manipulation (R), and window slicing (S).
\textbf{SCDA-WR}, \textbf{SCDA-WS}, and \textbf{SCDA-RS} represent that SCDA extracts fMRI features from any pair of perspectives, \ie, 
window warping (W) and receptive field manipulation (R), window warping (W) and window slicing (S), and receptive field manipulation (R) and window slicing (S), respectively. 
Taking SCDA-WR for instance, we construct a two-branch feature enrichment architecture and use fMRI features with window warping (W) and receptive field manipulation (R) as inputs.
Also, a mutual-consistency constraint is leveraged to reduce feature distribution gap across these two branches.

The average results of the SCDA and its variants in both cross-scanner and cross-study prediction tasks are reported in Fig.~\ref{fig_ablation_study}.  
For cross-scanner prediction, NYU$_1$ and UM are used as source and target domains, respectively, and the task is to differentiate ASD patients from HCs in UM. 
For cross-study prediction, NYU$_2$ is used as source domain and T2DM is used as target domain, and the task is to identify T2DM-CI patients from HCs in T2DM. 
% In cross-scanner prediction task, the largest site of ABIDE (\ie, NYU$_1$) is used as source domain and the second largest site of ABIDE (\ie, UM) is used as target domain, and the classification task is to differentiate ASD from HCs in UM.
% In cross-study prediction task, the largest site of REST-meta-MDD (\ie, Site-21) is used as source domain and the largest site of ABIDE (\ie, NYU$_1$) is used as target domain, and the classification task is to classify ASD from HCs in NYU$_1$.
From Fig.~\ref{fig_ablation_study}, we can see that SCDA is superior to SCDAw/oDA, and the underlying reason is that our SCDA considers cross-domain data heterogeneity by leveraging fMRI features inherent in the target domain, which helps improve target inference.
\emph{Besides}, the performance of SCDA is better than SCDAw/oS and SCDAw/oT.
This implies that using spatial and temporal attention improves diagnostic performance in SCDA, which can help the model focus on discriminative brain regions and time windows for prediction. % to final prediction.
\emph{Furthermore}, we can observe that SCDA consistently outperforms its degraded variants that capture fMRI representations from only two perspectives (\ie, SCDA-WR, SCDA-WS, and SCDA-RS). 
This indicates that our proposed multi-perspective feature enrichment strategy %, \eg, window warping, receptive field manipulation, and window slicing, 
could learn complementary features from different views, and 
the collaboration of these features can greatly improve cross-domain prediction. 
\if false
It is noted that we can easily add more branches in SCDA to exploit fMRI data from more perspectives. 
For instance, other feature enrichment techniques (\eg, magnitude warping~\cite{um2017dataaug}, jittering) can also be leveraged to further facilitate fMRI representation learning. 
\fi 

\subsection{Effect of Involved Databases in Unsupervised Pretraining}

\begin{table}[!t]
\setlength{\belowcaptionskip}{-2pt}
\setlength{\abovecaptionskip}{-2pt}
\setlength{\abovedisplayskip}{-2pt}
\setlength{\belowdisplayskip}{-2pt}
	\scriptsize
	\renewcommand\arraystretch{1}
	\caption{The results of the SCDA with different databases involved in unsupervised pretraining in classifying ASD from HC in cross-scanner prediction task. The `*' denotes the SCDA is statistically significantly different from a competing method.}\label{table_involved_databases}
	\centering
	\setlength\tabcolsep{1pt}
	\vspace{2mm}
        \resizebox{\linewidth}{!}{
	\begin{tabular*}{0.5\textwidth}{@{\extracolsep{\fill}} l|cccccc}
	\toprule
	~Method &  AUC (\%) & ACC (\%) & F1 (\%) & SEN (\%) & SPE (\%) & PRE (\%) \\
	\midrule
        SCDA-ABIDE$^*$ & 70.50±1.27 & 65.14±3.54 & 64.47±2.62 & 67.58±10.08 & 62.97±14.42 & 63.11±5.27 \\
        SCDA-MDD$^*$ & 71.02±1.65 & 64.86±2.41 & 62.82±6.03 & 65.45±16.33 & 64.32±16.03 & 63.67±5.32 \\
        SCDA-ADHD$^*$ & 69.18±2.49 & 64.57±2.98 & 64.17±2.09 & 67.27±4.13 & 62.16±7.40 & 61.66±3.67 \\
        SCDA & \textbf{72.79±2.10} & \textbf{68.00±0.83} & \textbf{66.77±2.74} & \textbf{68.79±7.94} & \textbf{67.30±6.53} & \textbf{65.48±1.92} \\
        \bottomrule
	\end{tabular*}
        }
\end{table}

% \begin{figure}[!t]\centering
% 	\setlength{\belowcaptionskip}{-0pt}
% 	\setlength{\abovecaptionskip}{-0pt}
% 	\setlength\abovedisplayskip{-8pt}
% 	\setlength\belowdisplayskip{-0pt}
% 	\includegraphics[width=0.48\textwidth]{figures/fig_involved_databases.pdf}
% 	\caption{AUC and ACC results of the proposed SCDA with different databases involved in unsupervised pretraining in classifying ASD from HC in cross-scanner prediction task.}\label{fig_involved_databases}
% \end{figure}

As mentioned in Section~\ref{sec_unsup_tr}, we propose an unsupervised pretraining strategy based on 3,806 subjects from three large-scale auxiliary fMRI databases (\ie, ABIDE, REST-meta-MDD, and ADHD-200) to facilitate more comprehensive fMRI feature extraction.
We now investigate the impact of involved databases in differentiating ASD patients from HCs in cross-scanner prediction tasks, with results shown in Table~\ref{table_involved_databases}.
The \textbf{SCDA-ABIDE}, \textbf{SCDA-MDD}, and \textbf{SCDA-ADHD} denote that only one specific database participates in unsupervised pretraining. 
%, while SCDA-ABIDE\&MDD, SCDA-ABIDE\&ADHD, and SCDA-MDD\&ADHD represent two databases involved in unsupervised pretraining.
Except for databases used in pretraining, all other settings of these methods remain the same. 

As shown in Table~\ref{table_involved_databases}, the SCDA pretrained with three databases generally outperforms its variants with one database involved for unsupervised pretraining. 
This implies that data scale and data diversity %(denoted by \eg, disease type, population demographics) 
play important roles in pretraining a general fMRI feature encoder to improve cross-domain prediction.
% Additionally, we find that SCDA-ABIDE generally shows better prediction results in ASD vs. HC classification than other two variants with one involved database. The reason may be that ...
It's interesting to explore more fMRI databases (\eg, Human Connectome Project\footnote{https://www.humanconnectome.org/}) to further increase data size and enhance data diversity, which will be our future work. % and thus facilitate more effective cross-domain knowledge transfer.

\subsection{Influence of Source Model Initialization}

\begin{table}[!t]
\setlength{\belowcaptionskip}{-2pt}
\setlength{\abovecaptionskip}{-2pt}
\setlength{\abovedisplayskip}{-2pt}
\setlength{\belowdisplayskip}{-2pt}
	\scriptsize
	\renewcommand\arraystretch{1}
	\caption{The results of our SCDA with different source model initialization for ASD vs. HC classification in cross-scanner prediction task. The `*' denotes the SCDA is statistically significantly different from a competing method.}\label{tab_src_model_initialization}
	\centering
	\setlength\tabcolsep{1pt}
	\vspace{2mm}
	\begin{tabular*}{0.49\textwidth}{@{\extracolsep{\fill}} l|cccccc}
	\toprule
	~Method &  AUC (\%) & ACC (\%) & F1 (\%) & SEN (\%) & SPE (\%) & PRE (\%) \\
	\midrule
        SCDA-W$^*$ & 71.37±0.85 & 65.43±2.10 & 63.57±4.38 & 65.15±11.46 & 65.68±10.93 & 63.54±3.72 \\
        SCDA-R$^*$ & 67.10±1.25 & 61.86±2.66 & 59.87±7.98 & 63.64±17.17 & 60.27±17.61 & 61.02±6.58 \\
        SCDA-S$^*$ & 69.25±2.10 & 63.29±4.18 & 61.49±5.16 & 63.64±13.99 & 62.97±17.69 & 62.22±6.08 \\
        SCDA & \textbf{72.79±2.10} & \textbf{68.00±0.83} & \textbf{66.77±2.74} & \textbf{68.79±7.94} & \textbf{67.30±6.53} & \textbf{65.48±1.92} \\
        \bottomrule
	\end{tabular*}
\end{table}

\begin{figure}[!t]
\centering
\setlength{\belowcaptionskip}{-2pt}
\setlength{\abovecaptionskip}{-2pt}
\setlength{\abovedisplayskip}{-2pt}
\setlength{\belowdisplayskip}{-2pt}
	\includegraphics[width=0.49\textwidth]{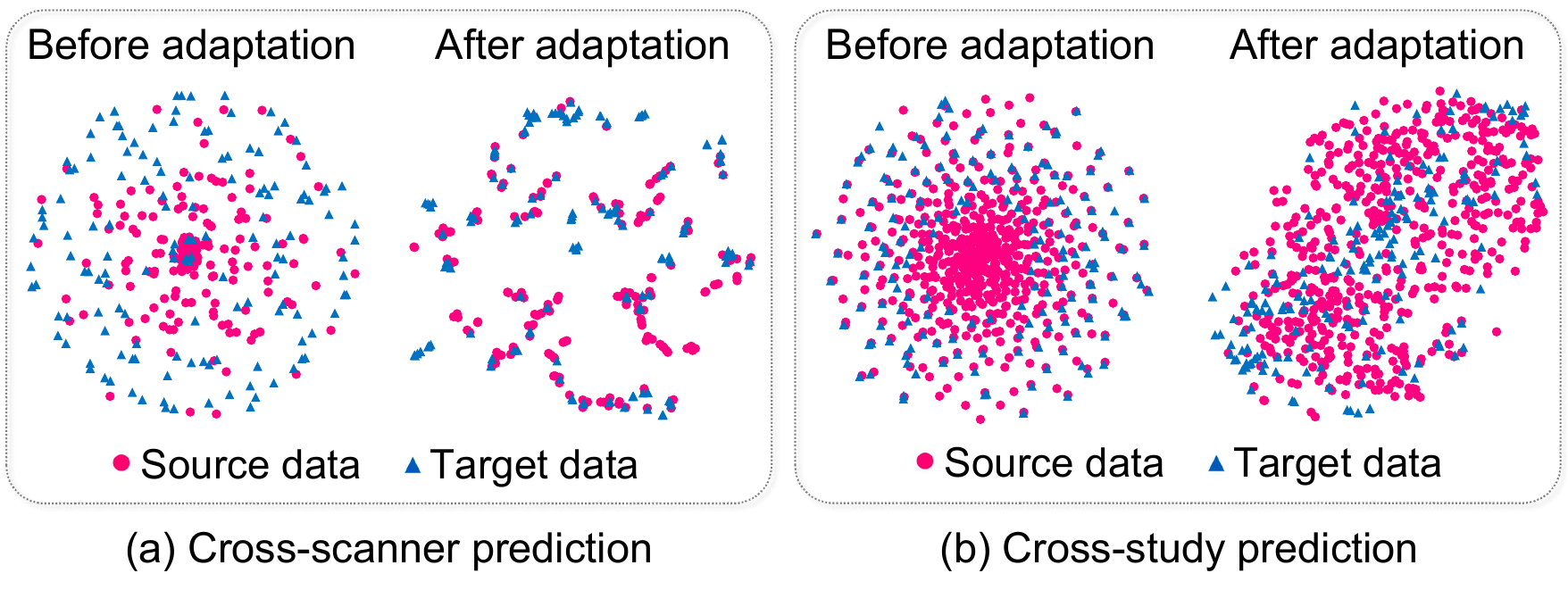}
	\caption{T-SNE visualization of feature distributions of source and target data before and after fMRI adaptation in (a) cross-scanner prediction task and (b) cross-study prediction task using the proposed SCDA. It can be seen that the cross-domain distribution gap is reduced after adaptation.}\label{fig_tsne}
\end{figure}

In this work, we first construct an initial MFE via unsupervised pretraining based on three auxiliary fMRI databases.
Then, we initialize the source model using average parameters of three branches of MFE.
In this section, to examine the influence of different initialized source models on classification performance, we use parameters of each branch of MFE to initialize the source model rather than the average of three branches. 
The results of our method and its three variants (called \textbf{SCDA-W}, \textbf{SCDA-R}, \textbf{SCDA-S}) in ASD vs. HC classification in the cross-scanner prediction task are shown in Table~\ref{tab_src_model_initialization}. 
Here, SCDA-W, SCDA-R, and SCDA-S denote that we initialize the source model using the branch with window warping (W), receptive field manipulation (R), and window slicing (S) for fMRI feature representation learning, respectively. 
% Besides, we also construct a multi-perspective source model (named SCDA-I-SSS) similar to the multi-perspective feature enrichment framework, which can be directly initialized by corresponding branches of the initial multi-perspective framework in unsupervised pretraining stage.
From Table~\ref{tab_src_model_initialization}, we can see that our SCDA consistently outperforms these competing methods.
This result is reasonable since averaged parameters from multiple branches can help extract more general fMRI features, which can facilitate more effective cross-domain knowledge transfer.
% This indicates that using averaged parameters from three branches in unsupervised pretraining leads to more generalized and informative fMRI features
% from databases with larger data scales and richer diversity
% This result is not surprising, since during unsupervised pretraining three branches are collaboratively learned via a mutual-consistency constraint, and many fMRI representations are shared among them.

% \subsection{Hyperparameters in Each Perspective}
% % each perspective's hyperparameter对结果的影响\\
% window slicing           {85\%, 90\%, 95\%, 100\%}\\
% window warping      {1., 1.1, 1.3, 1./1.1, 1./1.3}\\
% receptive field manipulation {40, 60, 80, 100}\\ 

\subsection{Visualization of fMRI Features}

We employ t-SNE~\cite{van2008visualizing} to visualize fMRI feature distributions before and after domain adaptation (via SCDA) in cross-scanner and cross-study prediction tasks, with results shown in Fig.~\ref{fig_tsne}.
In the cross-scanner prediction task, NYU$_1$ and UM are used as source and target domains, respectively. 
In the cross-study prediction task, Site-21 and NYU$_1$ serve as source and target domains, respectively. 
% In cross-scanner prediction task, the largest site of ABIDE (\ie, NYU$_1$) is used as source domain and the second largest site of ABIDE (\ie, UM) is used as target domain. 
% In cross-study prediction task, the largest site of REST-meta-MDD (\ie, Site-21) is used as source domain and the largest site of ABIDE (\ie, NYU$_1$) is used as target domain. 
In our study, each subject's fMRI timeseries is represented by an $L\times N$ matrix, with $L$ time points and $N$ brain ROIs.  
Before domain adaptation, we represent each subject using the flattened vector from the $L\times N$ matrix.
Then, we optimize SCDA using unlabeled target data and finally derive a feature encoder.
We use this feature encoder to capture fMRI feature representation, which is used to represent each subject after domain adaptation.
In Fig.~\ref{fig_tsne}, different colors represent different domains, \ie, magenta and blue denote source and target domains, respectively.
It can be seen from Fig.~\ref{fig_tsne} that after adaptation using our SCDA, features learned from source and target domains are mixed and mapped into a common latent feature space. 
It implies that our SCDA helps reduce the cross-domain data distribution gap.

\subsection{Identified Discriminative Brain ROIs}

\begin{figure}[!t]\centering
\setlength{\belowcaptionskip}{-2pt}
\setlength{\abovecaptionskip}{-2pt}
\setlength{\abovedisplayskip}{-2pt}
\setlength{\belowdisplayskip}{-2pt}
	\includegraphics[width=0.45\textwidth]{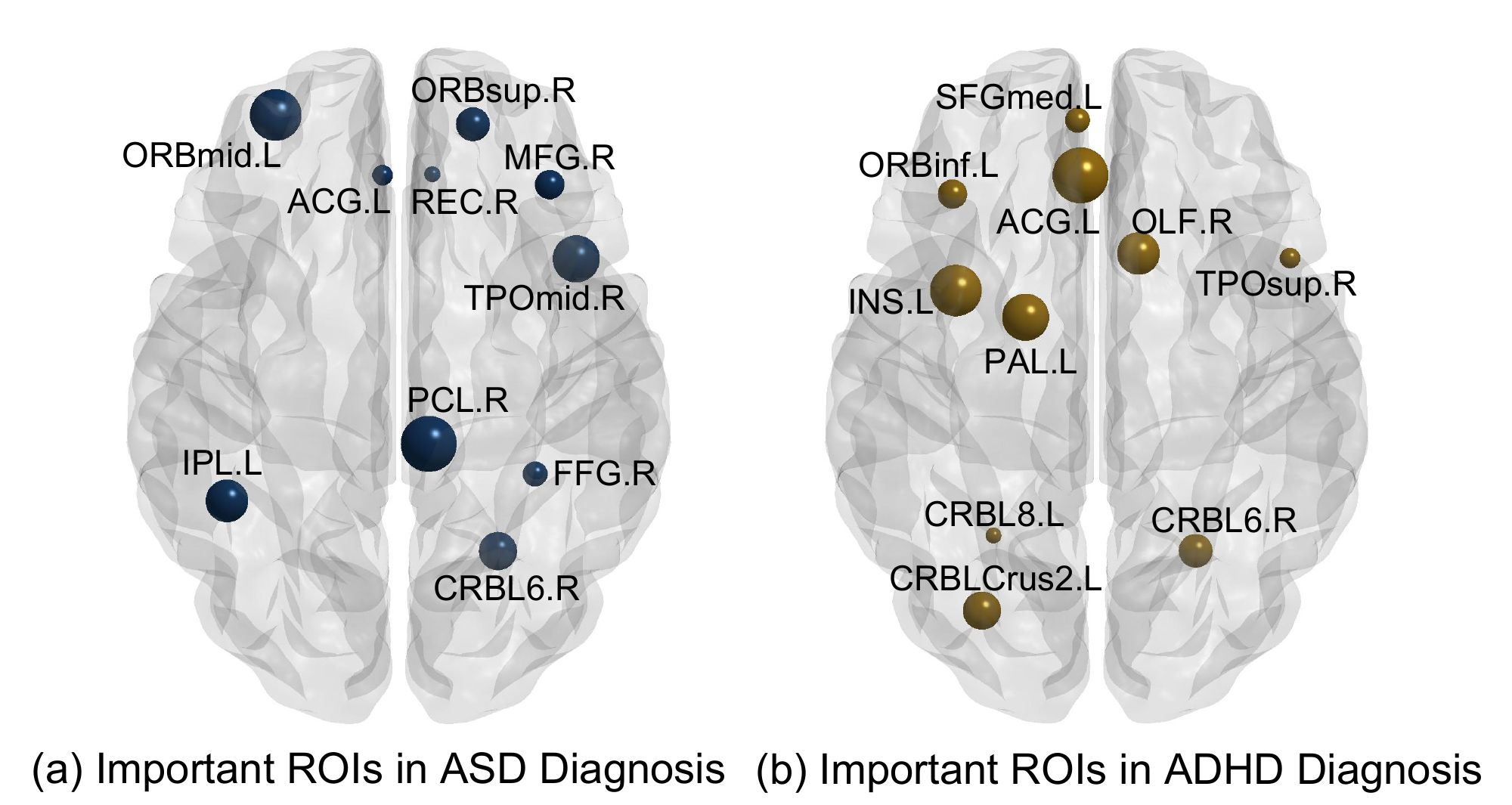}
    \caption{Top 10 important brain ROIs identified by our SCDA in two cross-scanner prediction tasks, \ie, (a) ASD diagnosis and (b) ADHD diagnosis.
    L: left hemisphere; R: right hemisphere.}\label{fig_discriminative_ROI}
\end{figure}

%As mentioned in Section~\ref{sec_feature_encoder}, 
We employ the squeeze-excitation mechanism~\cite{hu2018squeeze} to automatically learn spatial attention $M \in\mathbb{R}^{N}$ across different brain ROIs ($N$=116), with each element denoting the discriminative ability of the corresponding brain ROI for prediction. 
%That is, each subject is with a $116$-dimensional vector, where each element denotes the importance of the corresponding brain ROI in classification prediction.
We visualize the top 10 important brain ROIs identified by our SCDA in two cross-scanner prediction tasks in Fig.~\ref{fig_discriminative_ROI}~(a) and Fig.~\ref{fig_discriminative_ROI}~(b), respectively.
The first task is ASD diagnosis on the UM site and the second task is ADHD diagnosis on the Peking site. 
We show the averaged importance score of correctly-classified subjects with a disease, where a larger circle denotes more important ROIs, corresponding to a higher value.

From Fig.~\ref{fig_discriminative_ROI}~(a), we can see that \emph{right paracentral lobule} (PCL.R) plays the most important role in identifying ASD patients from HCs.
This finding is consistent with a previous study~\cite{libero2019longitudinal}, where this brain region is reported to exhibit consistently lower resting-state local connectivity in ASD subjects compared to typically developing controls. 
The reason may be that the paracentral lobule can operate with precuneus to produce a sense of self in a spatial world. 
Its impairment may disrupt the body and spatial representation in ASD patients and affect their development of theory of mind~\cite{cheng2015autism}. 
Moreover, functional abnormalities of our identified ROIs, \eg, middle frontal gyrus, orbital part (ORBmid.L), temporal pole: middle temporal gyrus (TPOmid.R), inferior parietal, supramarginal and angular gyri (IPL.L), middle frontal gyrus (MFG.R), have been verified in previous studies~\cite{xu2019altered, xu2020specific, wymbs2021altered, crider2014dysregulation}. 
These regions may be used as potential biomarkers to improve ASD diagnosis.

From Fig.~\ref{fig_discriminative_ROI}~(b), we find that \emph{anterior cingulate and paracingulate gyri} (ACG.L) plays the most important role in ADHD diagnosis, which is consistent with~\cite{makris2010anterior}.
The possible reason may be that ACG involves various brain functions, \eg, executive control and cognitive processing, and its abnormality may lead to the breakdown of attention allocation and cognition modulation, causing inattention and hyperactivity in ADHD patients~\cite{makris2010anterior}.
Besides, other identified ROIs by our SCDA, \eg, \emph{insula} (INS.L), \emph{lenticular nucleus, pallidum} (PAL.L), \emph{superior frontal gyrus, medial} (SFGmed.L), have also been verified associated with functional abnormalities in previous ADHD studies~\cite{gao2021structural, samea2019brain, jiang2019functional}.

% This is consistent with previous findings that functional coordination between bilateral homotopic brain ROIs is impaired in MDD patients, while the abnormalities in homotopic connectivity across hemispheres may imply abnormal neural circuits associated with aberrant emotional processing and cognition in MDD patients~\cite{xx}.
% Besides, Table~\ref{xx} suggests that our identified ROIs include lingual gyrus and calcarine fissure, while their abnormal FCs have been reported in adults with major depression~\cite{xx}.
% A potential reason could be that gray matter volume of STG changes after depression~\cite{xx}. 
% These findings further verify the reliability of the proposed UFA-Net in identifying disease-associated functional connectivity patterns for automated MDD identification with rs-fMRI data.

\subsection{Limitations and Future Work} 
There are several limitations in the current study. 
\emph{First}, our SCDA is a white-box source-free adaptation approach, where the parameters of the source model are accessible, which may suffer from data leakage problems under membership inference attack~\cite{nasr2019comprehensive}.
In the future, we will investigate source-free adaptation in black-box settings, where only prediction of the source model is available, which could further help protect data privacy in practical scenarios. 
\emph{Second}, this work explores three feature enrichment strategies to exploit fMRI representations for target model construction.
We will investigate more feature enrichment techniques (\eg, magnitude warping~\cite{um2017dataaug}) to further facilitate fMRI learning and enhance target inference.  
\emph{In addition}, our SCDA only leverages functional MRI for disease diagnosis without taking advantage of structural MRI data and/or non-imaging information (\eg, age, gender).
Future work will seek to combine complementary feature characteristics from different modalities to further improve learning performance.
\emph{Lastly}, current work learns brain functional connectivity features based on a single brain atlas (\ie, AAL) with 116 brain ROIs.
In the future, we will incorporate fMRI representations from multi-scale atlases for brain parcellation, %(\eg, Craddock with 200 ROIs~\cite{craddock2012whole}), 
which may provide different fine-grained information to further boost classification performance. 

% \emph{Lastly}, our SCDA investigates source-free domain adaptation by transferring source knowledge to one target domain only. Future work will explore multi-target adaptation scenarios.

\section{Conclusions}\label{sec_conclusion}
This paper introduces a source-free collaborative domain adaptation (SCDA) framework for fMRI-based neurological disorder diagnosis.
Our SCDA achieves source-to-target domain adaptation only based on a pretrained source model and unlabeled target data.
Specifically, a multi-perspective feature enrichment method is designed, which contains multiple collaborative branches to dynamically exploit target data from different views. 
The parameters of MFE initialized by the pretrained source model can be fine-tuned on unlabeled target fMRI, thus catering to data distributions of the target domain. 
Furthermore, we propose an unsupervised pretraining strategy to leverage large-scale auxiliary fMRIs from three public databases to further improve fMRI representation learning. 
Extensive experiments in cross-scanner and cross-study prediction tasks show the effectiveness of our proposed method. 
%And the model pretrained on large-scale rs-fMRI data has been released. 

% Future work will seek to exploit features from other modalities, different brain atlas, and investigate black-box domain adaptation scenarios.

\if false
\section*{Acknowledgments}
Y.~Fang, A. Bozoki, and M.~Liu were supported by NIH grant RF1AG073297. 
\fi 

\bibliographystyle{IEEEtran}
\bibliography{SFUDA_refs}

\end{document}